\documentclass{article}

\PassOptionsToPackage{numbers, compress}{natbib}

\usepackage[final]{neurips_2025}

\usepackage[utf8]{inputenc} % allow utf-8 input
\usepackage{amssymb}
\usepackage{pifont} % for check and cross marks 
\usepackage[T1]{fontenc}    % use 8-bit T1 fonts
\usepackage{hyperref}       % hyperlinks
\usepackage{url}            % simple URL typesetting
\usepackage{booktabs}       % professional-quality tables
\usepackage{amsfonts}       % blackboard math symbols
\usepackage{nicefrac}       % compact symbols for 1/2, etc.
\usepackage{microtype}      % microtypography
\usepackage{xcolor}         % colors
\usepackage{longtable}      % for long tables
\usepackage{amsmath}
\usepackage{nicefrac}       % compact symbols for 1/2, etc.
\usepackage{multirow}
\usepackage{enumerate}
\usepackage{multirow}
\usepackage{graphicx}
\usepackage{natbib}
\usepackage{float}
\usepackage{subcaption}
\usepackage{array}
\usepackage{neurips_2025}   % NIPS style file
\usepackage{algorithm}
\usepackage{algpseudocode}
\usepackage{subcaption}
\usepackage{makecell}
\usepackage{array}
\usepackage{pgfplots}
\usepackage{subcaption}
\usepackage{amsmath,amsfonts,bm}

\def\eqref#1{equation~\ref{#1}}
\def\1{\bm{1}}

\DeclareMathAlphabet{\mathsfit}{\encodingdefault}{\sfdefault}{m}{sl}
\SetMathAlphabet{\mathsfit}{bold}{\encodingdefault}{\sfdefault}{bx}{n}

\definecolor{olive}{RGB}{0.6, 0.6, 0.2}
\definecolor{sand}{RGB}{0.8666666666666667, 0.8, 0.4666666666666667}
\definecolor{wine}{RGB}{0.5333333333333333, 0.13333333333333333, 0.3333333333333333}
\definecolor{deblue}{RGB}{11,132,147}
\definecolor{ocra}{RGB}{204, 119, 34} % Custom commands, such as commands

\usepackage{latexml}
\iflatexml
\usepackage{authblk}
\fi

\title{Structured Temporal Causality for Interpretable Multivariate Time Series Anomaly Detection}

\iflatexml
\author{\textbf{Dongchan Cho}\textsuperscript{*}}
\affil{Industrial AI Lab, SimPlatform Co. Ltd. Affiliate Research Institute\\
  \href{mailto:cdc0213@simplatform.com}{\texttt{cdc0213@simplatform.com}}\\
  \textsuperscript{*}Corresponding author.}
\author{\textbf{Jiho Han}}
\affil{Industrial AI Lab, SimPlatform Co. Ltd. Affiliate Research Institute\\
  \href{mailto:jihohan@simplatform.com}{\texttt{jihohan@simplatform.com}}}
\author{\textbf{Keumyeong Kang}}
\affil{Industrial AI Lab, SimPlatform Co. Ltd. Affiliate Research Institute\\
  \href{mailto:kykang@simplatform.com}{\texttt{kykang@simplatform.com}}}
\author{\textbf{Minsang Kim}}
\affil{Industrial AI Lab, SimPlatform Co. Ltd. Affiliate Research Institute\\
  \href{mailto:mskim@simplatform.com}{\texttt{mskim@simplatform.com}}}
\author{\textbf{Honggyu Ryu}}
\affil{Industrial AI Lab, SimPlatform Co. Ltd. Affiliate Research Institute\\
  \href{mailto:hgryu@simplatform.com}{\texttt{hgryu@simplatform.com}}}
\author{\textbf{Namsoon Jung}}
\affil{Industrial AI Lab, SimPlatform Co. Ltd. Affiliate Research Institute\\
  \href{mailto:nsj@simplatform.com}{\texttt{nsj@simplatform.com}}}
\else
\author{%
  Dongchan Cho\thanks{Corresponding Author.} \qquad Jiho Han \qquad Keumyeong Kang \qquad \textbf{Minsang Kim} \\ \\ \textbf{Honggyu Ryu} \qquad \textbf{Namsoon Jung}\\ \\
  Industrial AI Lab, SimPlatform Co. Ltd. Affiliate Research Institute\\
  \texttt{\{cdc0213, jihohan, kykang, mskim, hgryu, nsj\}@simplatform.com}}
\fi

\begin{document}

% %%%%%%%%%%%%%%%%%
% % Main paper
% %%%%%%%%%%%%%%%%%
% \begin{center}
%     \color{red}
%     \large Provided proper attribution is provided, SimPlatform Co. Ltd. hereby grants permission to reproduce the tables and figures in this paper solely for use in journalistic or scholarly works.
% \end{center}

\maketitle
\begin{abstract}
Real-world multivariate time series anomalies are rare and often unlabeled. Additionally, prevailing methods rely on increasingly complex architectures tuned to benchmarks, detecting only fragments of anomalous segments and overstating performance. In this paper, we introduce OracleAD, a simple and interpretable unsupervised framework for multivariate time series anomaly detection. OracleAD encodes each variable’s past sequence into a single causal embedding to jointly predict the present time point and reconstruct the input window, effectively modeling temporal dynamics. These embeddings then undergo self-attention mechanism to project them into a shared latent space and capture spatial relationships. These relationships are not static, since they are modeled by a property that emerges from each variable's temporal dynamics. The projected embeddings are aligned to a Stable Latent Structure (SLS) representing normal-state relationships. Anomalies are identified using a dual scoring mechanism based on prediction error and deviation from the SLS, enabling fine-grained anomaly diagnosis at each time point and across individual variables. Since any noticeable SLS deviation originates from embeddings that violate the learned temporal causality of normal data, OracleAD directly pinpoints the root-cause variables at the embedding level. OracleAD achieves state-of-the-art results across multiple real-world datasets and evaluation protocols, while remaining interpretable through SLS.
\end{abstract}

\section{Introduction}
\label{introduction}
Unsupervised anomaly detection in multivariate time series (MTSAD)~\cite{ACM-Review-2021, ACM-Survey-2024} underpins reliability and safety in domains such as industrial control systems~\cite{SWaT-2016, Issues_Advances_AD-2018}, healthcare monitoring~\cite{ECG-2015, PhysioNet-2000}, and cyber-physical security~\cite{PSM-2021}. In these settings, anomalies are rare, unlabeled, and context-dependent, posing challenges to conventional methods and requiring actionable insights to maintain operational reliability. Models must therefore not only flag irregularities but also explain their causal origins. Thus, effective MTSAD methods should (i) define anomalies within a multivariate temporal framework, (ii) model the temporal and spatial signals of anomaly formation, and (iii) integrate these signals for both detection and root-cause diagnosis.

Although intrinsic temporal and structural changes occur when anomalies emerge, most MTSAD methods overlook these signals. They often rely on reconstruction errors computed over fixed "normal" and "abnormal" windows, on superficial feature-level differences after arbitrary model transformations, or on frequency-domain representations divorced from causal time dynamics. More complex deep-model pipelines (e.g., dual-axis attention) further amplify this issue by creating artificial separations between normal and anomalous data through their complex forward-propagation outputs, rather than by explicitly modeling how anomalies form and spread. Such complexity rarely translates into practical gains, as these methods frequently fail to outperform simpler baselines or to justify their architectural overhead empirically or theoretically~\cite{QuoVadisTAD-2024,LTSF-Linear-2023}. Transformer-based approaches exemplify this misalignment by using large temporal contexts and bidirectional attention, which ignore the unidirectional, irreversible nature of time and compromise real-time deployment. Furthermore, recent studies~\cite{TTMs-2024,LTSF-Linear-2023} demonstrate that compact causal architectures with lightweight designs can match or exceed Transformer performance across diverse forecasting benchmarks while greatly reducing computational costs.

At the core of these issues lies a fundamental misunderstanding regarding what naturally constitutes an anomaly in multivariate time series. We argue that anomalies inherently manifest through two interconnected signals. First, anomalies begin as prediction errors when a variable’s current state deviates from the expectations derived from its past states, signifying a breakdown in temporal causality within that variable. Second, temporal disruptions propagate into structural deviations by breaking the stable inter-variable relationships that are typically preserved under normal conditions. Thus, we explicitly define anomalies in multivariate time series as processes originating from a loss of temporal causality in specific variables, which subsequently alters the inter-variable relational structure.

Motivated by these perspectives, we propose \textbf{OracleAD}, an interpretable framework that captures temporal causality and leverages the natural signals inherent in multivariate anomalies. OracleAD processes the time series data with a sliding window of length \(L\). For each variable \(i\), a per-variable LSTM encoder processes its past \(L-1\) observations to produce hidden states \(\{h^1_{i}, \dots, h^{L-1}_{i}\}\). We then apply attention pooling over these hidden states to yield a single \emph{causal embedding} \(c_i\), which summarizes the temporal information necessary to predict the final timepoint \(x^L_i\) in the window. To instill temporal causality in the encoder, we jointly train it with two objectives: one-step point prediction of \(x^L_i\) and reconstruction of the window. The learned causal embeddings are projected into a shared latent space via a self-attention mechanism, where attention weights quantify dynamic interaction strengths between variables at each time point.

During training, OracleAD computes pairwise dissimilarities between projected causal embeddings across variables at each time step, forming a dissimilarity matrix that explicitly reflects the current spatial organization. At the end of each training epoch, these matrices are aggregated into a \textit{Stable Latent Structure (SLS)}, which serves as a stable, statistically derived reference structure representing consistent inter-variable relationships under normal conditions. At inference time, OracleAD employs a dual scoring system. The \textit{prediction score} quantifies the error between the actual and predicted states at each time step, signaling temporal irregularities that violate past-informed expectations. The \textit{deviation score} captures structural inconsistencies by measuring the deviation between the current latent dissimilarity matrix and the SLS, flagging shifts in inter-variable relationships that deviate from learned causal structures.
By integrating these complementary signals, OracleAD not only detects anomalies accurately but also transparently pinpoints root-cause variables at the embedding level, addressing critical limitations of existing methods and providing causal explanations of anomaly formation and propagation.

OracleAD demonstrates state-of-the-art anomaly detection performance across multiple real-world datasets. Experimental results show consistent improvements in detection accuracy, anomaly localization, and robustness under diverse evaluation protocols. Recognizing and explicitly modeling natural anomaly signals enables OracleAD to detect anomalies more efficiently using compact temporal windows. This approach is supported by recent findings~\cite{AnomalyTransformer-2021}, which show that anomalies strongly correlate with adjacent time points. OracleAD provides interpretability through a dual scoring system based on prediction error and structural deviation, enabling users to identify anomalies and trace their root causes at both the temporal and variable levels. By grounding anomaly detection in temporal causality and stable latent representations, OracleAD addresses core limitations of existing methods and offers a principled, practical direction for advancing multivariate time series anomaly detection.

\section{Background}
\label{background:MTSAD}

\subsection{Unsupervised Multivariate Time Series Anomaly Detection}

Early approaches to multivariate time-series anomaly detection primarily relied on classical outlier detection methods \cite{LOF-2000F,Isolation_Forest-2008, PCA-2003}. However, these early methods treated each time point as an independent multivariate sample and did not leverage the rich temporal dependencies inherent in sequential data. Subsequently, research gradually shifted toward deep sequence modeling based on reconstruction or prediction. EncDec-AD~\cite{EncDec-AD-2016}, OmniAnomaly~\cite{OmniAnomaly-2019}, and recurrent ensemble variants~\cite{RecurrentAE-2019} train autoencoder or variational models to reproduce normal behavior, flagging large reconstruction or prediction errors as anomalies. Although these approaches are conceptually simple and robust, they treat each channel independently, failing to capture inter-variable dependencies critical for complex fault patterns. MAD-GAN~\cite{MADGAN-2019} extends this line by applying adversarial training via LSTM-based generator-discriminator networks, enabling joint modeling of temporal dynamics and inter-variable structures. However, unstable GAN optimization often limits robustness and interpretability.

To address this, explicit spatio-temporal modeling techniques emerged. MSCRED~\cite{MSCRED-2019} uses ConvLSTM~\cite{ConvLSTM-2015} to reconstruct correlation matrices over sliding windows, and InterFusion~\cite{InterFusion-2021} employs hierarchical VAEs to disentangle temporal and spatial factors. Transformer-based architectures inject attention mechanisms into MTSAD. Anomaly Transformer~\cite{AnomalyTransformer-2021} characterizes anomalies as discrepancies between learned attention weights and prior temporal associations, whereas TranAD~\cite{TranAD-2022} combines predictive and adversarial objectives to sharpen sensitivity to subtle deviations. Prior to this, Gangopadhyay et al.~\cite{STattention-2021} proposed a spatio-temporal attention mechanism for interpretive forecasting in MTS, showing that attention can offer explanatory cues across time and variable dimensions. PatchTST~\cite{PatchTST-2022} and SparseTSF~\cite{SparseTSF-2024} further improve efficiency by processing fixed-size patches or periodic samples, but at the expense of strict causal consistency and fine-grained localization. Graph-wise and dimension-wise attention methods capture variable interactions more flexibly~\cite{GraphAtt-2020}. GDN~\cite{GDN-2021} learns a data-driven adjacency matrix to model dependencies (albeit static at inference), while Crossformer~\cite{Crossformer-2022} and iTransformer~\cite{iTransformer-2023} deploy multi-axis attention to fuse temporal and cross-variable signals, though large receptive fields may dilute local anomalies. Recently, Zheng et al.~\cite{STgraph-2023} introduced a correlation-aware spatial-temporal graph learning model. Contrastive and spectral approaches offer orthogonal gains: DCdetector~\cite{DCdetector-2023} forgoes reconstruction entirely, using contrastive grouping of normal versus perturbed views; SARAD~\cite{SARAD-2024} adds regularization across adjacent subsequences; CATCH~\cite{CATCH-2025} isolates spectral anomalies via frequency-domain patching.

\subsection{Limitations of Existing Benchmarks}
\label{background:benchmarks}
Several widely used MTSAD benchmarks, such as SWaT~\cite{SWaT-2016}, SMAP, and MSL~\cite{SMAP_MSL-2018}, are frequently regarded as fully multivariate datasets. However, in practice, most real-world attacks and faults affect only a limited number of variables. For example, many scenarios in the SWaT dataset involve perturbations to only a small subset of sensors. Similarly, in the SMAP and MSL datasets, many sensor streams exhibit behavior that closely resembles step functions. A recent study by Liu et al.~\cite{Elephant_in_the_Room-2024} demonstrates that converting these datasets into univariate formats results in comparable detection performance. This observation raises concerns regarding the actual necessity of using complex models that are specifically designed to capture cross-variable dependencies.

In addition to dataset limitations, the evaluation protocols used in MTSAD research exhibit several inconsistencies~\cite{TSADEvaulation-2022, TSAD-illusion-2021}. The point-adjusted F1 score, for instance, tends to give excessive credit for any detection that falls within an anomaly window, even if the localization is inaccurate. Metrics such as AUC-ROC and AUC-PR are threshold-independent and widely used, but they often produce misleading results in highly imbalanced settings where true negatives dominate. The Affiliation F1 score~\cite{Local_Evaluation_TSAD-2022} attempts to compensate for localization errors through post-processing, but it does not fundamentally resolve the problem of poor anomaly localization. More recently, researchers have proposed volume-under-surface metrics such as VUS-ROC and VUS-PR~\cite{Elephant_in_the_Room-2024, VUS-2022}, which jointly measure detection consistency and temporal precision across a range of thresholds.

Considering these limitations, it is difficult to regard strong performance on a single benchmark or metric as a reliable indicator of general robustness~\cite{NNContribute-2022}. In order to provide an accurate assessment of MTSAD methods, it is necessary to conduct comprehensive evaluations that include multiple anomaly types, realistic variable correlations, and diverse metrics such as VUS-based measures.

\section{Method}

\begin{figure}
    \centering
    \includegraphics[width=1\linewidth]{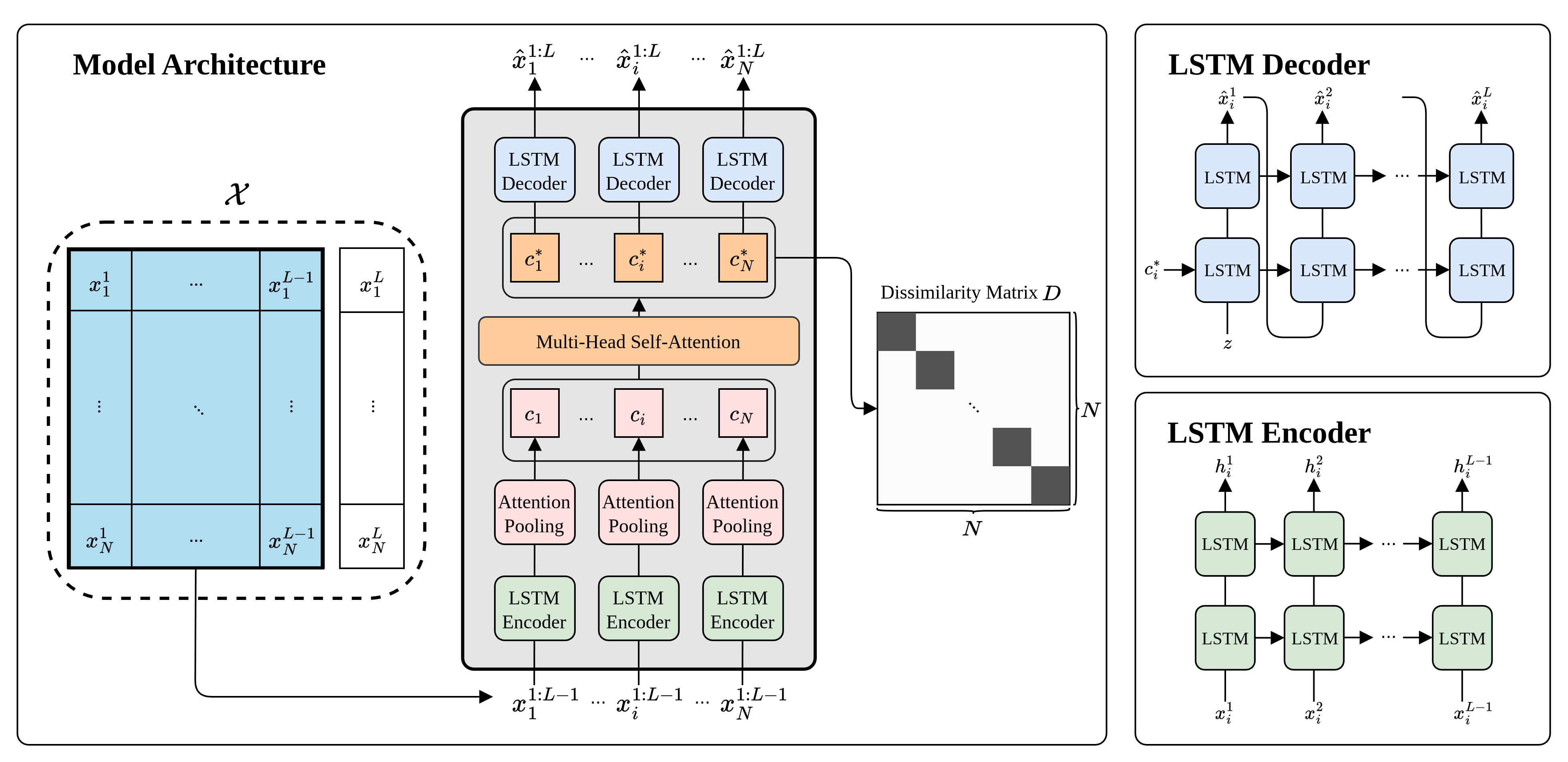}
    \caption{Overview of the OracleAD pipeline. Each variable’s past window is encoded by an LSTM and pooled via attention. The resulting embeddings are refined through multi-head self-attention to capture inter-variable dependencies. These embeddings are then fed into an LSTM decoder for reconstruction and next-step prediction. Finally, the pairwise distance matrix of the embeddings is compared against the Stable Latent Structure (SLS) to detect anomalies. The decoder uses an all-zero vector $z$ as its initial input.}
    \label{fig:pipeline}
\end{figure}

Let $\mathcal{X} \in \mathbb{R}^{N \times L}$ denote a fixed-length input window of multivariate observations, defined as

\begin{equation}
\mathcal{X} = \{\mathbf{x}^1, \mathbf{x}^2, \dots, \mathbf{x}^L\},
\quad \mathbf{x}^\ell \in \mathbb{R}^N,
\label{eq:window_def}
\end{equation}

where each window element is written as

\begin{equation}
\mathbf{x}^\ell = (x_1^\ell, x_2^\ell, \dots, x_N^\ell)^\top,
\quad 1 \leq \ell \leq L,
\label{eq:window_element}
\end{equation}

and $\mathbf{x}^L$ is the most recent observation vector.  
OracleAD determines whether $\mathbf{x}^L$ is anomalous based on (i) its prediction error relative to temporal patterns learned within the window, 
and (ii) its deviation from the stable inter-variable structure observed under normal conditions.

\subsection{Temporal Causality Modeling}

OracleAD models each variable independently using a per-variable encoder--decoder architecture. This contrasts with prior approaches~\cite{TimesNet-2022, AnomalyTransformer-2021, MSCRED-2019} that process the multivariate sequence jointly, often with a shared model. However, in real-world settings, different variables often follow heterogeneous dynamics, and anomalies may affect only a subset of them. Shared architectures can entangle unrelated temporal patterns, hindering both detection sensitivity and interpretability. To address this, we adopt a modular structure where each variable learns its own causal dynamics.

As shown in Figure~\ref{fig:pipeline}, for each variable segment 
$\mathbf{x}_i = (x_i^1, x_i^2, \dots, x_i^{L-1})^\top \in \mathbb{R}^{L-1}$, 
an LSTM~\cite{EncDec-AD-2016} encoder $\text{Enc}_i$ processes the input window and outputs a sequence of hidden states:

\begin{equation}
\{h_i^1, h_i^2, \dots, h_i^{L-1}\}, \quad h_i^l \in \mathbb{R}^d,
\label{eq:lstm_outputs}
\end{equation}

where $d$ is the hidden dimension.
To distill salient temporal information and suppress noise, we apply a learnable attention mechanism over the hidden states:

\begin{equation}
f(h_i^l) = w^\top h_i^l + b, \quad w \in \mathbb{R}^d, \; b \in \mathbb{R},
\label{eq:scalar_score}
\end{equation}

followed by a softmax normalization over the temporal axis ($l=1,\dots,L-1$):

\begin{equation}
\alpha_i^l = \text{softmax}(f(h_i^l)), 
\quad 
c_i = \sum_{l=1}^{L-1} \alpha_i^l\, h_i^l.
\label{eq:temporal_attention}
\end{equation}

The resulting vector $c_i \in \mathbb{R}^d$, termed the causal embedding of variable $i$, aggregates its past hidden states weighted by their relevance in predicting $x_i^L$, thereby encoding the temporal causality inherent in that variable's history.

To incorporate inter-variable interactions, we apply Multi-Head Self-Attention (MHSA)~\cite{Transformer-2017} to the stacked embeddings

\begin{equation}
C = [c_1,\dots,c_N]^\top \in \mathbb{R}^{N\times d},
\label{eq:embedding_matrix}
\end{equation}

yielding the attention-refined output

\begin{equation}
C^* = [c_1^*,\dots,c_N^*] \in \mathbb{R}^{N\times d}.
\label{eq:mha_output}
\end{equation}

Each $c_i^*$ is computed as

\begin{equation}
c_i^* = \sum_{h=1}^{H} W_h^O \,
\mathrm{softmax}\!\Bigl(\tfrac{(W_h^Q c_i)(W_h^K C)^\top}{\sqrt{d_h}}\Bigr)\,
W_h^V C,
\label{eq:mha}
\end{equation}

where $H$ denotes the number of attention heads, and 
$W_h^Q, W_h^K, W_h^V, W_h^O$ are learnable projection matrices for query $Q$, key $K$, value $V$, and output in head $h$. 
The per-variable embeddings \(\{c_1,\dots,c_N\}\) are first projected into a shared latent space via head-specific linear transformations. Subsequent multi-head self-attention mixes these projections, so each \(c_i^{*}\) absorbs contextual cues from all other variables and captures soft, dynamic dependencies without a predefined static graph.

These context-aware embeddings $c_i^*$ are then passed into their corresponding LSTM decoders $\text{Dec}_i$, each trained to both reconstruct the past window and predict the next value:

\begin{equation}
\hat{x}_i^{1:L-1}, \; \hat{x}_i^{L} = \text{Dec}_i(c_i^*).
\label{eq:decoder}
\end{equation}

The outputs of each decoder are used to compute the reconstruction loss
$\|x_i^{1:L-1} - \hat{x}_i^{1:L-1}\|^2$
and the prediction loss
$\|x_i^{L} - \hat{x}_i^{L}\|^2$,
which together ensure that each causal embedding $c_i$ faithfully captures its variable’s temporal dynamics under a causal, windowed regime.

The final multivariate prediction at time $L$ is assembled as

\begin{equation}
\hat{\mathbf{x}}^L = [\hat{x}_1^L, \hat{x}_2^L, \dots, \hat{x}_N^L] \in \mathbb{R}^N.
\label{eq:final_prediction}
\end{equation}

This per-variable framework guarantees robust modeling of the normal data distribution by (i) condensing all $L-1$ past steps into each causal embedding $c_i$, thereby covering the full spectrum of normal temporal patterns rather than overfitting to local snippets, (ii) employing independent encoder--decoder branches so that each $c_i$ adapts to its variable’s unique dynamics, (iii) fusing these embeddings via MHSA into context-aware vectors $c_i^*$, which learn the joint normal manifold and soft correlations across variables, and (iv) suppressing the influence of rare spikes through attention pooling and Stable Latent Structure (SLS)-based structural regularization.

\subsection{Stable Latent Structure (SLS)}
\label{section:sls}
To detect structural anomalies and promote latent consistency during training, OracleAD maintains a reference structure called the \textbf{Stable Latent Structure (SLS)}. The SLS summarizes stable inter-variable relationships observed in the latent space during normal behavior. Unlike static graphs or manually defined priors, it is constructed directly from data using the attention-refined latent representations of individual variables.

Let $c_i^{*} \in \mathbb{R}^{d}$ denote the latent vector of variable $i$ ($i=1,\dots,N$) after temporal encoding and relational attention. For each training window $k$, we form a pairwise dissimilarity matrix

\begin{equation}
D^{(k)}_{ij} = \bigl\| c_i^{*(k)} - c_j^{*(k)} \bigr\|_2,
\quad D^{(k)} \in \mathbb{R}^{N \times N},
\label{eq:dissimilarity}
\end{equation}

where we adopt the L2 distance because it accounts for both directional mismatch and latent energy differences in the embeddings. 
Empirically, L2 also provided more stable and discriminative results than alternatives such as cosine similarity or L1 distance (see Appendix~\ref{appendix:distance_comparison}).

Aggregating over the $M$ windows that appear in one training epoch yields the stable template

\begin{equation}
\mathbf{SLS} = \frac{1}{M} \sum_{k=1}^{M} D^{(k)},
\label{eq:sls}
\end{equation}

which provides a compact baseline of canonical inter-variable structure. Since the first epoch has no prior SLS to build upon, the updating process begins at its end, and the resulting SLS is then used from the second epoch onward. During training, it acts as a regularizer that encourages the encoders to preserve this relational pattern. At inference time, for each time step $t$ in the entire input sequence, OracleAD computes the deviation matrix

\begin{equation}
\mathcal{D}^t_\text{matrix} = \big| D^t - \mathbf{SLS} \big|,
\label{eq:deviation_matrix}
\end{equation}

which measures element-wise deviations from the normal inter-variable structure. 
High-magnitude rows or columns in $\mathcal{D}^t_\text{matrix}$ indicate variables that disrupt relational stability, enabling root-cause identification.

\subsection{Training Objective}
\label{section:training_objective}

OracleAD is trained with a composite loss function that enforces temporal causality, reconstructs normal patterns, and regularizes inter-variable structure. The total loss is defined as

\begin{equation}
\mathcal{L} =
\underbrace{\|\,\mathbf{x}^L - \hat{\mathbf{x}}^L\,\|^2}_{\text{prediction loss}}
+ \lambda_{\mathrm{recon}} \cdot 
\underbrace{\|\,\mathbf{x}^{1:L-1} - \hat{\mathbf{x}}^{1:L-1}\,\|^2}_{\text{reconstruction loss}}
+ \lambda_{\mathrm{dev}} \cdot 
\underbrace{\frac{1}{N^2} \sum_{i=1}^N \sum_{j=1}^N (D_{ij} - \mathbf{SLS}_{ij})^2}_{\text{deviation loss}},
\label{eq:total_loss}
\end{equation}

where $\mathbf{x}^\ell \in \mathbb{R}^N$ denotes the $\ell$-th element of the window and $D$ is the dissimilarity matrix defined in Eq.~\ref{eq:dissimilarity}.

The first term encourages each decoder to forecast the most recent observation $\mathbf{x}^L$ using only the preceding window elements, thereby enforcing temporal causality.  
The second term guides the model to reconstruct the past window $\mathbf{x}^{1:L-1}$, preserving contextual memory and improving robustness to noise. This reconstruction term prevents the model from relying on shortcut predictions based only on local trends, forcing it instead to leverage the full temporal context within the window.  
The third term penalizes discrepancies between the current dissimilarity matrix $D$ and the reference SLS from Eq.~\ref{eq:sls}, ensuring that latent representations remain aligned with relational patterns observed under normal conditions.  

The hyperparameters $\lambda_{\mathrm{recon}}$ and $\lambda_{\mathrm{dev}}$ (set to $0.1$ and $3$ by default, respectively) balance the reconstruction and structural regularization terms against the prediction objective, promoting both predictive accuracy and structural coherence. In the first epoch, the deviation loss is omitted because no SLS is available; it is included from the second epoch onward once the initial SLS has been constructed.

\subsection{Anomaly Scoring}
\label{section:anomaly_scoring}

During inference, OracleAD computes two complementary scores for each time step $t$ in the entire input sequence.  
The prediction score is defined as

\begin{equation}
\mathcal{P}^t_\text{score} = \frac{1}{N}\sum_{i=1}^N \bigl|\,x_i^t - \hat{x}_i^t\,\bigr|,
\label{eq:pred_score}
\end{equation}

which measures the average absolute error between the observed variables and their reconstructions.  

The deviation score is defined as

\begin{equation}
\mathcal{D}^t_\text{score} = \bigl\|\, D^t - \mathbf{SLS}\,\bigr\|_F,
\label{eq:dev_score}
\end{equation}

where $D^t$ is the pairwise dissimilarity matrix of the causal–context embeddings at time $t$, and $\mathbf{SLS}$ is the stable relational template from Eq.~\ref{eq:sls}. Here $\|\cdot\|_F$ denotes the Frobenius norm. 

The final anomaly score is then defined as

\begin{equation}
\mathcal{A}^t_\text{score} = \mathcal{P}^t_\text{score} \cdot \mathcal{D}^t_\text{score},
\label{eq:anom_score}
\end{equation}

which jointly captures both temporal and spatial anomalies in multivariate time series. 
By grounding detection in both temporal prediction and relational stability in latent space, OracleAD achieves enhanced sensitivity and interpretability compared to methods relying solely on univariate errors or static dependency graphs. 
The diagnostic role of each component is illustrated with real-world examples in Section~\ref{experiment:diagnosis}.\

\section{Experiments}
\subsection{Experimental Setup}

\paragraph{Datasets.}
We evaluate OracleAD on three widely adopted benchmark datasets: \textbf{SMD}~\cite{OmniAnomaly-2019}, \textbf{PSM}~\cite{PSM-2021}, and \textbf{SWaT}~\cite{SWaT-2016}. These datasets span diverse industrial scenarios, including cloud server telemetry (SMD, 38 variables), real-world industrial sensors (PSM, 25 variables), and a water treatment testbed (SWaT, 51 variables). Each dataset contains labeled anomaly segments with varying temporal and spatial characteristics. Full details are provided in Appendix~\ref{appendix:datasets}.
\paragraph{Evaluation Metrics.}
To comprehensively evaluate model performance, we report results using seven metrics: standard F1 (not point-adjusted), range-based F1 (R-F1)~\cite{PrecRec_for_TS-2018}, affiliation F1 (Aff-F1)~\cite{Local_Evaluation_TSAD-2022}, AUC-ROC (A-ROC)~\cite{ROC-2006}, AUC-PR (A-PR)~\cite{PR_ROC-2006}, VUS-ROC (V-ROC), and VUS-PR (V-PR)~\cite{VUS-2022}. These metrics capture different aspects of anomaly detection, including temporal accuracy, interval-level coherence, robustness to class imbalance, and threshold stability. More detailed descriptions of each metric are provided in Appendix~\ref{appendix:metrics}.
\paragraph{Baselines.}
\label{sec: baselines}
We benchmark OracleAD against a diverse set of classical and deep learning baselines. For reconstruction-based method, \textbf{AutoEncoder (A.E})~\cite{AD_AE_NDR-2014} and \textbf{OmniAnomaly (Omni)}~\cite{OmniAnomaly-2019} are considered. Spatially-structured and attention-based deep models include \textbf{AnomalyTransformer (A.T)}~\cite{AnomalyTransformer-2021}, \textbf{DCdetector (DC)}~\cite{DCdetector-2023}, \textbf{SARAD}~\cite{SARAD-2024} and \textbf{CATCH}~\cite{CATCH-2025}. Forecasting-oriented approaches such as \textbf{PatchTST (Patch)}~\cite{PatchTST-2022}, \textbf{TimesNet (TsNet)}~\cite{TimesNet-2022}, \textbf{DLinear (DLin)}~\cite{LTSF-Linear-2023}, \textbf{NLinear (NLin)}~\cite{LTSF-Linear-2023},  \textbf{iTransformer (iTrans)}~\cite{iTransformer-2023} and \textbf{ModernTCN (Modern)}~\cite{ModernTCN-2023} are also considered, due to their high performance in general time-series tasks.  
All baselines are run using publicly available implementations and reimplementations based on the original paper's configurations. The implementation details of baselines are provided in Appendix~\ref{appendix:baselines}.
\paragraph{Implementation Details}
\label{sec: implemantation details}
OracleAD is implemented in a fully unsupervised setting using only normal training data. We use a sliding window of length $L = 10$ for all datasets, and exclude the first $L - 1$ time steps of the test data from evaluation. Following the recommendations from recent benchmark studies~\cite{Unsupervised_Model_Selection-2022, Elephant_in_the_Room-2024, TSAD_comprehensive_eval-2022}, we consider thresholding to be orthogonal to the quality of the model itself. Therefore, we either report threshold-independent metrics or evaluate detection performance under an optimal threshold to better isolate the effectiveness of the anomaly scoring function. Further details, sensitivity analysis of hyperparameters, and computational cost are provided in Appendix~\ref{appendix:implementation}.
 % Each variable is modeled using a 2-layer LSTM encoder and decoder with 256 hidden dimensions. Then We use a multi-head self-attention (MHSA) module with 8 heads and a 256-dimensional embedding to model inter-variable interactions. 

\begin{table}
\caption{Performance (\%) of deep anomaly detection models across seven evaluation metrics on three benchmark datasets (PSM, SMD, SWaT). All results are averaged over five random seeds. \textbf{Boldface} indicates the best score and \underline{underline} indicates the second-best score.}
\label{tab:main_results}
\centering
\setlength{\tabcolsep}{3pt}
\renewcommand{\arraystretch}{0.95}
\resizebox{\textwidth}{!}{
\begin{tabular}{c|c|cccclcccccccc}
\toprule
Dataset & Metric & A.E&Omni & A.T&Patch &TsNet& DLin& NLin& DC & iTrans & Modern & SARAD & CATCH & OracleAD \\
\midrule
& F1 & 47.55&\underline{45.90}& 43.45&43.45 &43.45& 43.45& 43.45 & 43.45& 43.45& 43.45& 45.75& 44.33& \textbf{65.85}\\
& R-F1 & 45.93&40.58& 12.17&34.75 &37.49& 34.74& 32.10& 21.35& 35.91& 35.48& 13.54& \underline{54.19}& \textbf{54.66}\\
& Aff-F1 & 73.70&73.49& 69.68&69.43 &69.43& 69.43& 69.43& 69.59& 69.53& 69.43& 77.12& \textbf{79.16}& \underline{78.07}\\
PSM & A-ROC & \underline{66.79}&63.95& 38.35&58.68 &59.09& 58.02& 58.56& 49.86& 59.22& 59.21& 62.86& 64.75& \textbf{84.78}\\
& A-PR & \underline{47.27}&45.00& 24.28&37.91 &38.95& 37.17& 37.64& 27.78& 38.27& 38.59& 41.67& 43.40& \textbf{68.11}\\
& V-ROC & \underline{68.20}&61.23& 50.96&50.62 &50.67& 50.53& 50.69& 50.11& 51.23& 50.68& 57.63& 57.00& \textbf{84.24}\\
& V-PR & 49.66&\underline{52.49}& 49.76&50.31 &49.86& 49.93& 46.71& 27.89& 44.54& 50.55& 38.64& 45.95& \textbf{68.17}\\
\midrule
& F1 & 25.78&\underline{32.16}& 7.98&7.98 &7.98& 7.98& 7.98 & 8.00& 7.98& 7.98& 25.92& 7.98& \textbf{43.03}\\
& R-F1 & 28.09&\underline{35.81}& 5.84&13.81 &9.43& 13.37& 13.51& 8.23& 9.02& 14.43& 10.35& 2.90& \textbf{38.88}\\
& Aff-F1 & \underline{83.34}&82.57& 67.43&67.43 &67.43& 67.43& 67.43 & 67.49& 67.43& 67.43& 75.87& 67.43& \textbf{84.73}\\
SMD & A-ROC & 76.80&71.25& 50.40&73.85 &59.09& 72.73& 73.83& 49.97& 74.57& 72.13& 72.84& \underline{80.96}& \textbf{83.56}\\
& A-PR & 19.40&\underline{27.73}& 4.57&14.82 &13.66& 13.88& 14.10& 4.16& 14.57& 13.01& 25.87& 17.09& \textbf{44.83}\\
& V-ROC & \textbf{74.38}&\underline{73.26}& 50.34&51.66 &51.10& 51.52& 51.76& 49.92& 51.76& 51.55& 65.57& 50.95& 69.57\\
& V-PR & 22.50&31.18& 36.86&\underline{41.60} &41.21& 41.26& 39.15& 4.27& 32.85& 41.54& 19.33& 35.25& \textbf{47.52}\\
\midrule
& F1 & 74.46&\underline{75.40}& 21.65&21.65 &21.65& 7.98 & 12.14& 21.65& 21.65& 21.65& 57.30 & 21.65& \textbf{76.50}\\
& R-F1 & \underline{36.63}&\textbf{38.72}& 15.48&12.23 &20.80& 5.13 & 12.06& 15.54& 11.89& 12.88& 21.37 & 14.74& 28.15\\
& Aff-F1 & \textbf{75.30}&72.09& 70.06&69.51 &73.47& 67.43 & 69.68& 69.25& 69.25& 72.08& \underline{74.88} & 73.68& 71.97\\
SWaT & A-ROC & 81.97&82.15& 42.58&24.36 &28.97& 50.48 & 23.31& 49.89& 24.57& 24.50& \textbf{85.40} & 33.31& \underline{82.71}\\
& A-PR & 67.51&\textbf{72.73}& 11.93&8.40 &10.71& 4.53 & 8.00& 12.14& 8.33& 9.00& 64.77 & 13.39& \underline{72.39}\\
& V-ROC & 81.89&81.98& 50.72&49.84 &20.80& 50.36 & 49.02& 50.00& 49.90& 50.15& \textbf{86.30} & 51.89& \underline{82.42}\\
& V-PR & \underline{65.89}&64.42& 17.00&11.92 &51.20& 28.47 & 10.55& 11.92& 12.05& 13.05& 62.72 & 18.70& \textbf{74.16}\\
\bottomrule
\end{tabular}
}
\end{table}

\subsection{Main Results}
\label{main results}

As shown in Table~\ref{tab:main_results}, OracleAD achieves the highest scores across most of the reported metrics on SMD (average over all subsets), PSM, and SWaT. This demonstrates strong robustness against diverse anomaly types and dimension of data. 

Among the various metrics, we emphasize the standard point-wise F1 score, which directly reflects alignment between detected anomalies and ground truth. OracleAD consistently outperforms all baselines with large margins in F1 (PSM:+19.95\%pt, SMD:+10.87\%pt, SWaT:+0.9\%pt), confirming its ability to clearly separate anomalous from normal points. These high scores are often accompanied by strong performance in threshold-independent metrics (AUC, VUS). Recent benchmarks~\cite{Elephant_in_the_Room-2024} identify VUS-PR as a key indicator of detection consistency and localization quality. In this metric, OracleAD significantly outperforms all baselines (PSM:+15.68\%pt, SMD:+5.92\%pt, SWaT:+8.27\%pt), which underscores its superior anomaly localization.

In contrast, Affiliation F1 tends to yield high scores across most models regardless of their performance on other metrics. This reveals a key limitation of the metric, Affiliation  F1 rewards predictions made near the anomaly segment, even if they do not directly overlap with the ground truth. Experiments in~\cite{Elephant_in_the_Room-2024} demonstrate that models can score highly on affiliation even with random predictions or by detecting only a small part of an anomaly segment. This explains why some baselines (e.g., A.T, CATCH) exhibit high affiliation F1 despite poor F1, AUC-PR, and VUS-PR, indicating their failure to fully capture anomaly regions. Overall, these results confirm OracleAD’s dual-score design delivers both sharp detection and robust localization across diverse scenarios.

\subsection{Visual Interpretation and Anomaly Diagnosis}
\label{experiment:diagnosis}

\begin{figure}[H]
\centering

\begin{subfigure}[b]{0.48\textwidth}
    \centering
    \includegraphics[width=\linewidth]{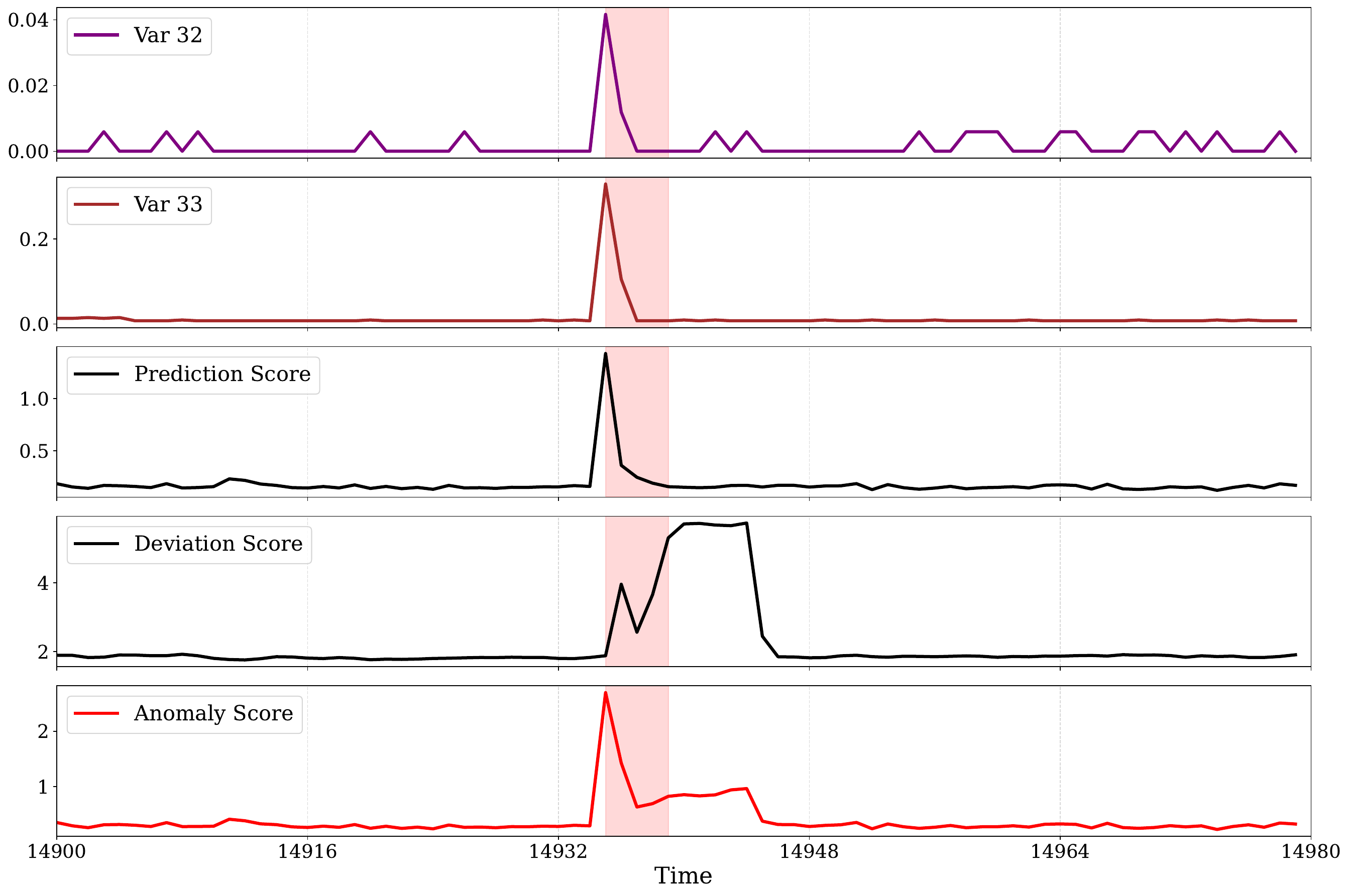}
    \caption{anomaly range=[14935, 14938]}
    \label{fig:plots-a}
\end{subfigure}
\hfill
\begin{subfigure}[b]{0.48\textwidth}
    \centering
    \includegraphics[width=\linewidth]{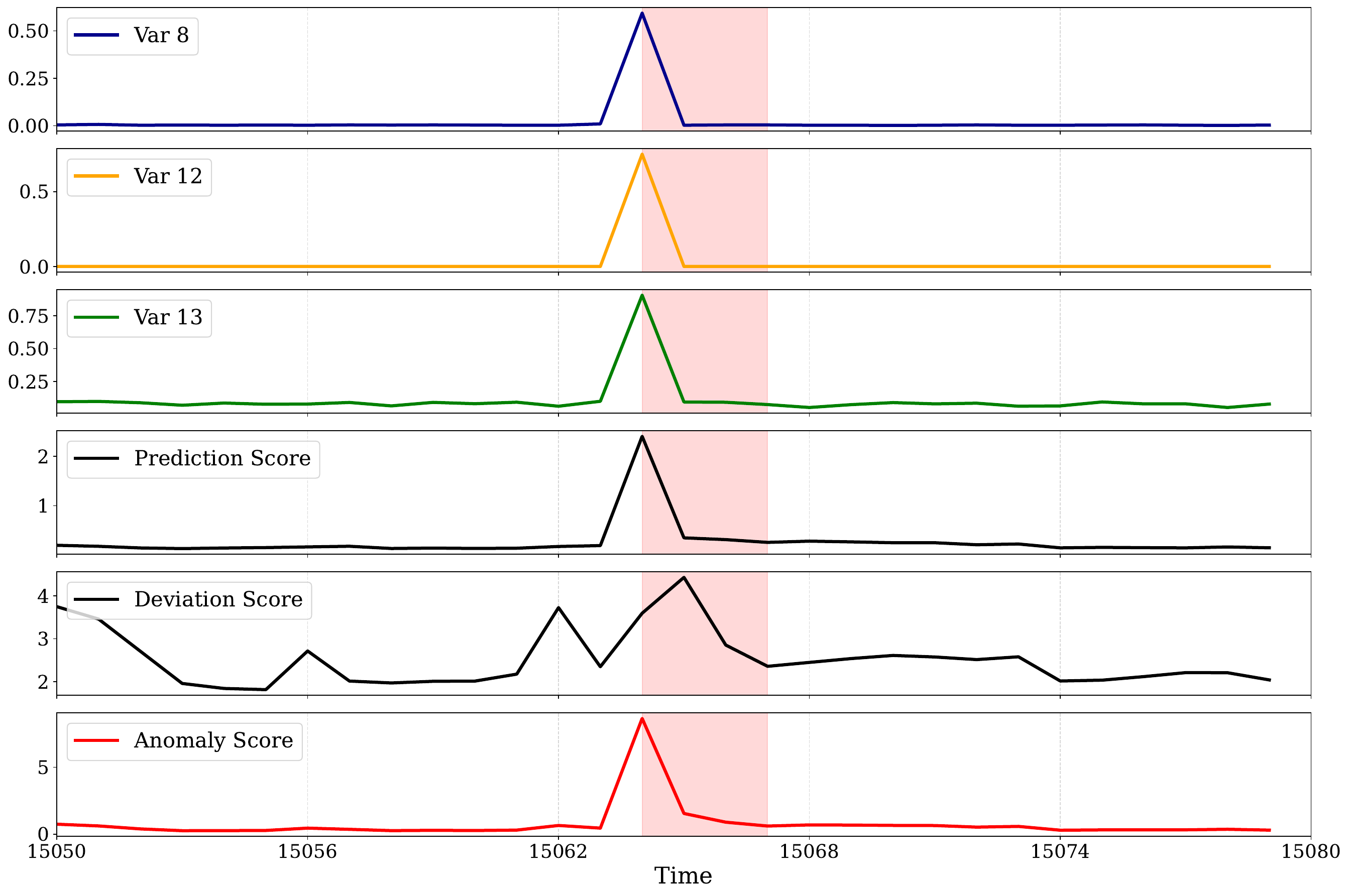}
    \caption{anomaly range=[15064, 15066]}
    \label{fig:plots-b}
\end{subfigure}

\vspace{1em}

\begin{subfigure}[b]{0.48\textwidth}
    \centering
    \includegraphics[width=\linewidth]{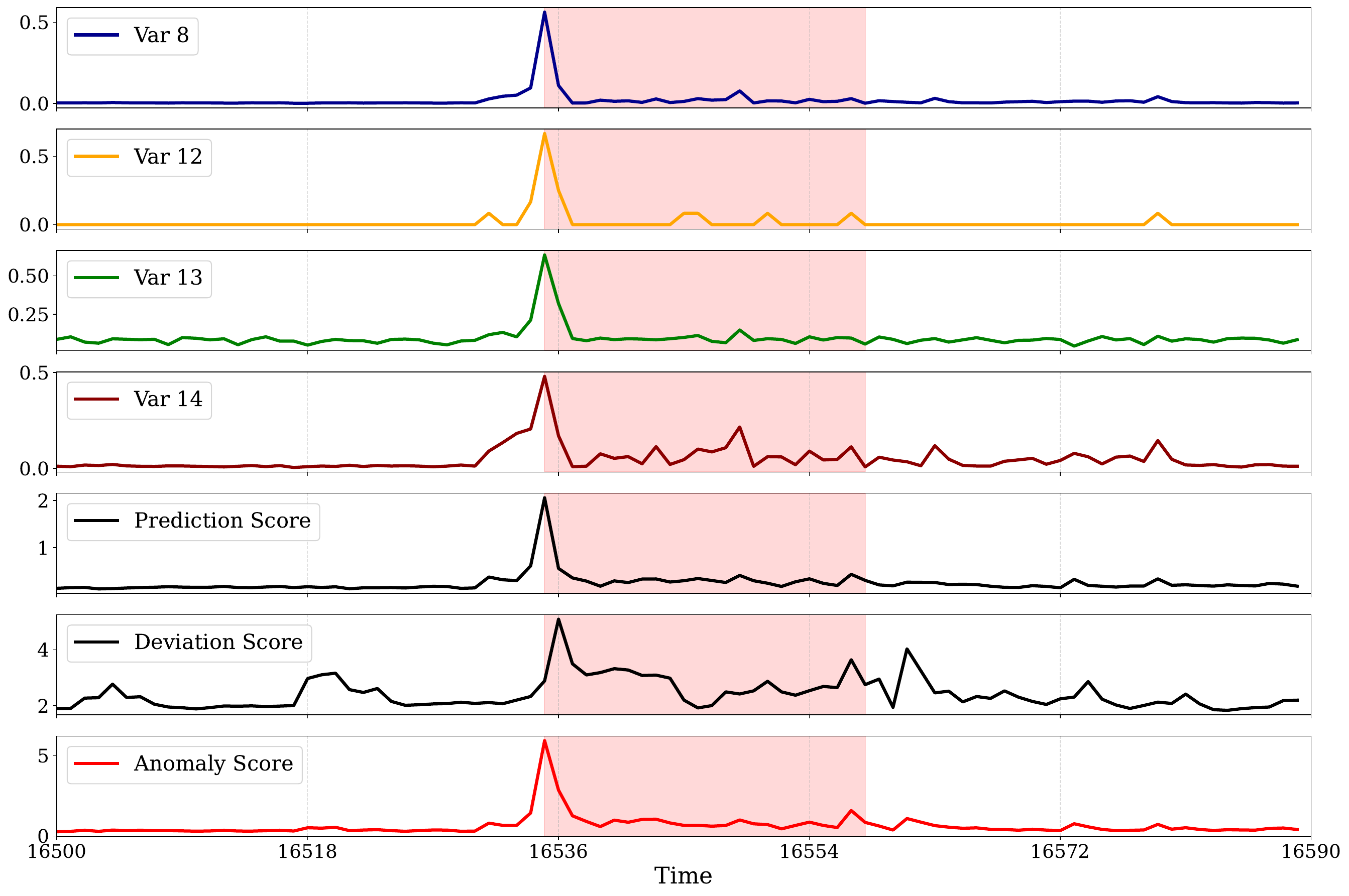}
    \caption{anomaly range=[16535, 16557]}
    \label{fig:plots-c}
\end{subfigure}
\hfill
\begin{subfigure}[b]{0.48\textwidth}
    \centering
    \includegraphics[width=\linewidth]{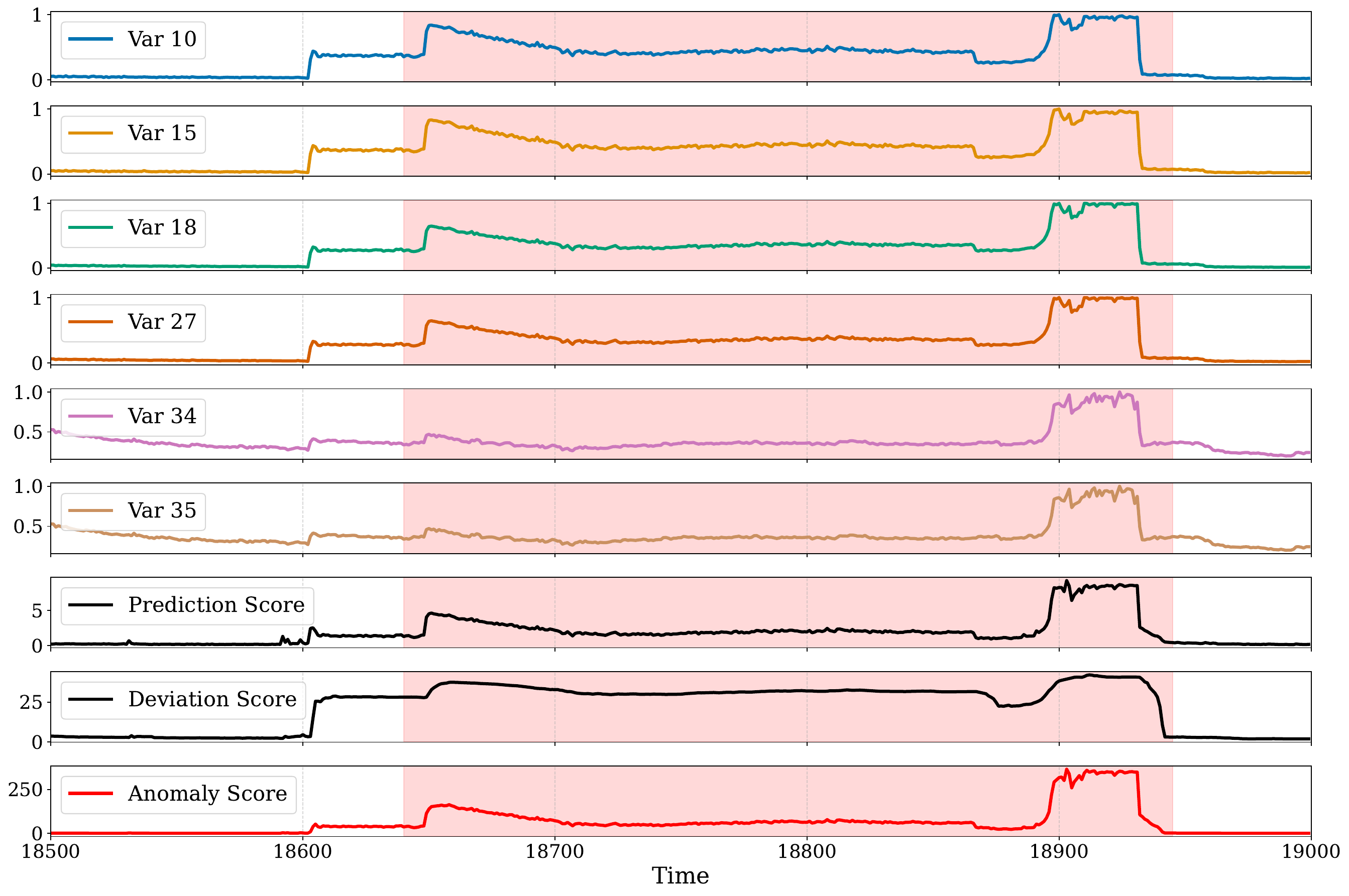}
    \caption{anomaly range=[18640, 18944]}
    \label{fig:plots-d}
\end{subfigure}

\caption{Raw signals of the identified root-cause variables and anomaly score visualization for selected intervals in the SMD dataset. Shaded red regions indicate ground-truth anomalies.}
\label{fig:plots}
\end{figure}

\begin{figure}[H]
  \begin{subfigure}[b]{0.24\textwidth}
    \includegraphics[width=\linewidth]{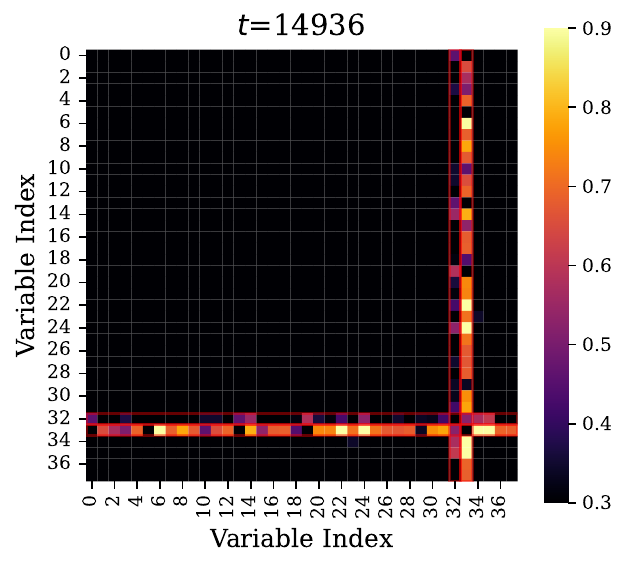}
    \caption{$\mathcal{D}^\text{14936}_\text{matrix}$}
    \label{fig:t=14936}
  \end{subfigure}
  \begin{subfigure}[b]{0.24\textwidth}
    \includegraphics[width=\linewidth]{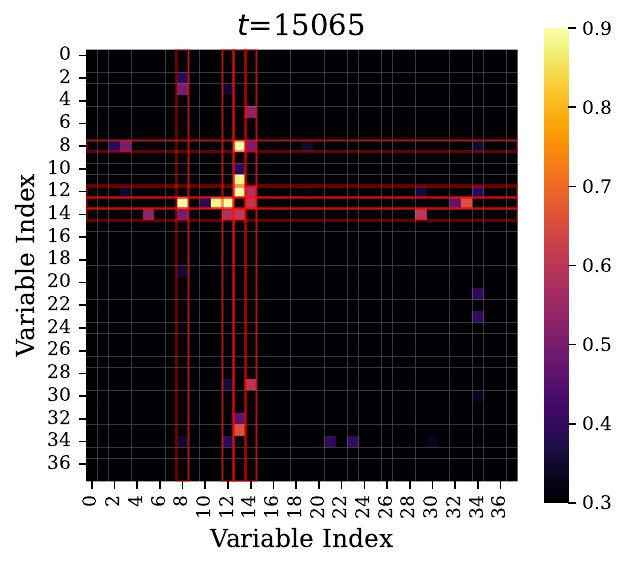}
    \caption{$\mathcal{D}^\text{15065}_\text{matrix}$}
    \label{fig:t=15065}
  \end{subfigure}
  \begin{subfigure}[b]{0.24\textwidth}
    \includegraphics[width=\linewidth]{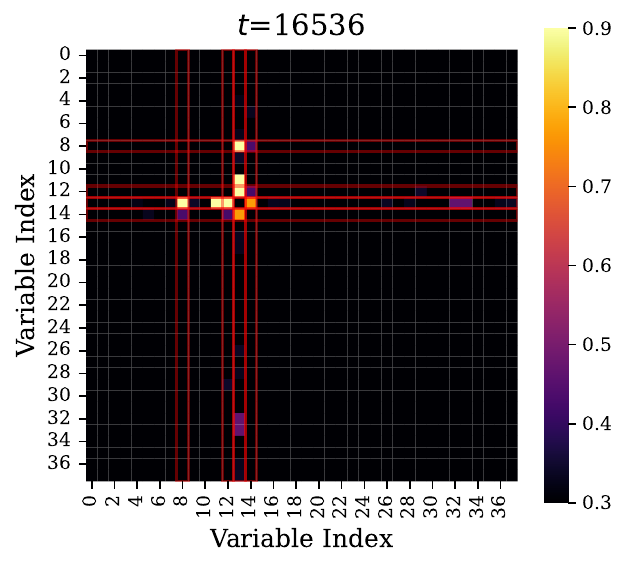}
    \caption{$\mathcal{D}^\text{16536}_\text{matrix}$}
    \label{fig:t=16536}
  \end{subfigure}
  \begin{subfigure}[b]{0.24\textwidth}
    \includegraphics[width=\linewidth]{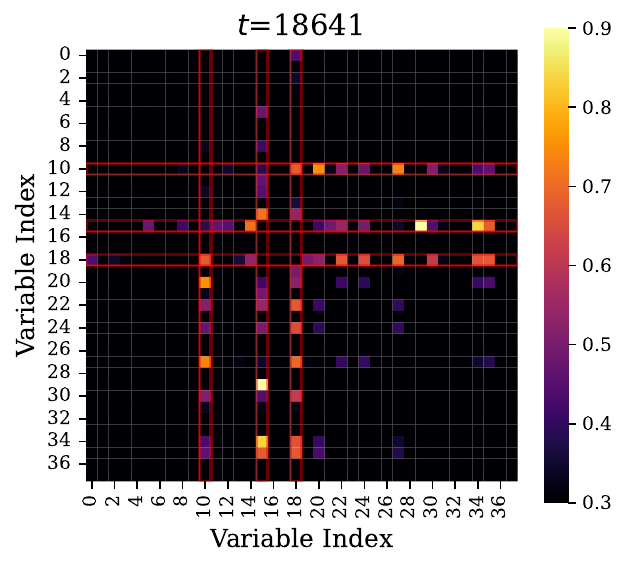}
    \caption{$\mathcal{D}^\text{18641}_\text{matrix}$}
    \label{fig:t=18641}
  \end{subfigure}
  \caption{Visualization for Deviation Matrices $\mathcal{D}^t_\text{matrix}$ at anomalous time points in SMD dataset. Each matrix corresponds respectively to the subfigures in Figure~\ref{fig:plots}. The color bar of each heatmap represents the magnitude of pairwise dissimilarity between variables. A bigger row over column value means a higher chance of an anomaly in that variable. For visualization, each matrix is independently min--max normalized and values are clipped to the range [0.3, 0.9] to enhance contrast.}  
  \label{fig:heatmaps}
\end{figure}

A key interpretability component is the deviation matrix $\mathcal{D}^t_\text{matrix} = |D^t - \mathbf{SLS}|$, which highlights each variable’s disruption of the learned structure. Figure~\ref{fig:heatmaps} visualizes deviation matrices at representative anomaly timestamps from the SMD dataset. For example, variables 32, 33 in Figure~\ref{fig:heatmaps}(\subref{fig:t=14936}) and 10, 15, 18 in Figure~\ref{fig:heatmaps}(\subref{fig:t=18641}) show dominant deviations. Their corresponding raw signals and scoring trends in Figure~\ref{fig:plots} validate these candidates, as each shows strong alignment with prediction peaks and anomaly score elevations, confirming OracleAD’s ability to localize root causes through unsupervised latent analysis. A practical heuristic is that variables with multiple highlighted rows or columns are more likely to be root causes, as their relational patterns remain consistently perturbed across time. When arranged sequentially, these matrices reveal how the structural disruptions of key variables evolve over the course of an anomaly segment. Quantitative analysis based on this row-wise aggregation and more examples are provided in Appendix~\ref{appendix:quantitative analysis}. 

\vspace{-1em}
\vspace{-1em}

As shown in Figure~\ref{fig:plots}, OracleAD combines prediction and deviation scores into a final anomaly score. 
The prediction score measures point-level discrepancies between observation and forecast, responding sharply to sudden deviations and achieving high temporal precision. 
However, its response is typically brief and localized to the moment of disruption, potentially missing prolonged structural anomalies.
In contrast, the deviation score compares the current dissimilarity matrix to the Stable Latent Structure (SLS) and captures longer-term relational disturbances (see Figure~\ref{fig:plots}(\subref{fig:plots-d})) that persist even after the raw signal stabilizes. 
This property aligns with real-world anomaly labeling, where fault intervals span the full duration of a system disturbance rather than just its initiation. 
Consequently, the deviation score remains elevated across an entire fault segment, reflecting sustained relational instability.

Nonetheless, the deviation score may exhibit a slight temporal lag because structural disruption can only be captured after anomalous patterns have entered the input window. This means the effect of a recent anomaly is only reflected after it enters the window, and the score can also be influenced by residual effects from recent anomalies (see Figure~\ref{fig:plots}(\subref{fig:plots-a})), potentially leading to false positives in recovery periods. The multiplicative combination of these two scores enables OracleAD to balance their respective strengths and mitigate their individual weaknesses. While the prediction score anchors detection to the precise moment of temporal disruption, the deviation score extends sensitivity across the whole anomaly segment.
This joint scoring mechanism also suppresses false positives and recovers false negatives. High deviation scores that arise in non-anomalous contexts are suppressed by low prediction error, while false negatives from prediction-only detection are compensated by the sustained elevation of structural deviations. Together, they offer both sharpness and stability, enabling OracleAD to detect anomalies with improved precision, temporal coverage, and interpretability across diverse anomaly types and durations. More case studies are provided in appendix \ref{appendix:case studies}.

\subsection{Ablation Study}
\vspace{-1em}
\label{sec:ablation}
% \begin{table}[H]
%   \centering
%   \scriptsize
%   \caption{Ablation study results (\%) of OracleAD variants on two benchmark datasets (PSM and SWaT) across F1, A-PR, and V-PR metrics. All results are averaged over five random seeds.}
%   \label{tab:ablation_loss}
%   \setlength{\tabcolsep}{3.7pt}
%   \begin{tabular}{c  |c |ccccc|ccccc}
%     \toprule
%     \multirow{2}{*}{\textbf{Component}} & \multirow{2}{*}{\textbf{Variant}} 
%       & \multicolumn{5}{c}{\textbf{PSM}}& \multicolumn{5}{c}{\textbf{SWaT}} \\
%     \cmidrule(lr){3-7} \cmidrule(lr){8-12} \
%     & & F1 &  A-ROC&A-PR &  V-ROC&V-PR & F1 &  A-ROC&A-PR &  V-ROC&V-PR \\
%     \midrule
%     \shortstack[l]{Loss Function(Sec~\ref{section:training_objective})}& w/o Reconstruction 
%       & 58.03&  75.77&54.62&  74.56&54.40& 76.61&  83.13&72.18&  82.92&71.95\\
%     \midrule
%     \multirow{2}{*}{\shortstack[l]{Scoring Strategy(Sec~\ref{section:anomaly_scoring})}} 
%       &  $\mathcal A_{score}$ = $\mathcal D_{score}$
%       & 59.06&  75.90&55.75&  75.78&56.11& 76.92&  81.61&71.71&  81.39&70.77\\
%     & $\mathcal A_{score}$ = $\mathcal P_{score}$
%       & 55.33&  77.54&60.98&  60.99&60.99& 70.49&  82.73&71.25&  82.68&68.50\\
%     \midrule
%  \multicolumn{2}{c|}{\textbf{OracleAD (Full Model)}}& 65.85 &  84.78& 68.11 &  84.24&68.17 & 76.50 &  82.71&72.39 &  82.42&74.16 \\
%  \bottomrule
%   \end{tabular}
% \end{table}

\begin{table}[H]
\caption{Ablation study results (\%) on the PSM, SMD, and SWaT datasets. All results are averaged over five random seeds.}
\label{tab:ablation}
\scriptsize
\makebox[\textwidth][c]{  
\setlength{\tabcolsep}{3.7pt}
\begin{tabular}{c|c|ccc|ccc|ccc}
  \toprule
  \multirow{2}{*}{\textbf{Component}} & \multirow{2}{*}{\textbf{Variant}} 
     &  \multicolumn{3}{c|}{\textbf{PSM}} &  \multicolumn{3}{c|}{\textbf{SMD}} &  \multicolumn{3}{c}{\textbf{SWaT}} \\
  \cmidrule(lr){3-11}
   & &  F1 & V-ROC & V-PR &  F1 & V-ROC & V-PR &  F1 & V-ROC & V-PR \\
  \midrule
  Loss Function(Sec~\ref{section:training_objective}) & w/o Reconstruction Loss
   & 58.03 & 74.56 & 54.40
   & 56.47 & 75.59 & 54.29
   & 76.61 & 82.92 & 71.95 \\
  \cmidrule(lr){1-11}
  \multirow{2}{*}{Anomaly Score(Sec~\ref{section:anomaly_scoring})} & $\mathcal{A}^t_\text{score}$ = $\mathcal{D}^t_\text{score}$
   & 59.06 & 75.78 & 56.11
   & 47.32 & 77.50 & 37.02
   & 76.92 & 81.39 & 70.77 \\
   & $\mathcal{A}^t_\text{score}$ = $\mathcal{P}^t_\text{score}$
   & 55.33 & 60.99 & 60.99
   & 58.98 & 87.39 & 53.71
   & 70.49 & 82.68 & 68.50 \\
  \cmidrule(lr){1-11} % 
  \multicolumn{2}{c|}{\textbf{OracleAD (Full Model)}} 
   & 65.85 & 84.24 & 68.17
   & 60.19 & 80.81 & 56.63
   & 76.50 & 82.42 & 74.16 \\
  \bottomrule
\end{tabular}
}
\end{table}

\vspace{-1em}
We conduct an ablation study to evaluate the contributions of reconstruction loss and anomaly scoring strategies. Results on PSM, SWaT, and SMD (average of 4 subsets) are summarized in Table~\ref{tab:ablation}. Per-subset results of SMD are provided in Appendix~\ref{appendix:experiments}.  

\textbf{Loss Function.}  
The full model achieves the highest performance across most of the datasets and metrics. Removing the reconstruction loss results in moderate degradation, especially in PSM (F1:-7.82\%pt, V-ROC:-9.68\%pt, V-PR:-13.77\%pt) and SMD (F1:-3.72\%pt, V-ROC:-5.22\%pt, V-PR:-2.34\%pt), indicating that reconstruction helps stabilize temporal representations and improve robustness.  

\textbf{Anomaly Score.}  
While deviation-only scoring ($\mathcal{A}^t_\text{score} = \mathcal{D}^t_\text{score}$) performs well on SWaT, it underperforms on PSM and SMD, likely because it lacks temporal context. Prediction-only scoring $(\mathcal{A}^t_\text{score} = \mathcal{P}^t_\text{score}$) can detect sharp temporal deviations but fails to capture relational disruptions, leading to weak performance on SWaT (F1:-6.01\%pt, V-PR:-6.66\%pt). As discussed in Section~\ref{background:benchmarks}, SWaT anomalies often affect only a small subset of variables, which can limit the effectiveness of prediction-based scoring. However, structural disruptions caused by these localized anomalies still alter inter-variable relationships, allowing deviation-based scoring to retain strong performance. The combined scoring ($\mathcal{A}^t_\text{score}=\mathcal{P}^t_\text{score}\cdot\mathcal{D}^t_\text{score}$) consistently performs best overall, confirming the complementarity of temporal and structural dimensions in OracleAD’s design.

\section{Conclusion}
\label{sec:conclusion}
We introduced \textbf{OracleAD}, an unsupervised framework that jointly models temporal causality and inter-variable structure through a unified latent representation. OracleAD derives a compact causal embedding for each variable, aligns them via attention into a shared latent space, and detects anomalies through deviations from a learned Stable Latent Structure (SLS). 

OracleAD achieves state-of-the-art performance across multiple real-world datasets and evaluation protocols, demonstrating strong robustness, localization accuracy, and interpretability. Its dual scoring mechanism captures both prediction errors and structural inconsistencies, enabling precise detection and fine-grained root-cause analysis even without labels. We also found limitations in our framework when applied to complex or multimodal systems, where globally consistent inter-variable relationships and continuous input assumptions may not hold. 

Our paper highlights the importance of grounding anomaly detection in temporal causality and latent relational consistency, establishing a principled foundation for generalizable and interpretable multivariate time series analysis.

% %%%%%%%%%%%%%%%%%
% % Bibliography  
% %%%%%%%%%%%%%%%%%
\newpage
\bibliographystyle{plainnat}
\bibliography{references}

%%%%%%%%%%%%%%%%
% Appendix
%%%%%%%%%%%%%%%%

\newpage
\appendix
\input{appendices/title.tex}
\section{Datasets}
\label{appendix:datasets}

\subsection{Dataset Statistics}
\label{appendix:datasets-statistics}
\begin{table}[h]
\scriptsize
\centering
\caption{Statistics of the benchmark datasets used for multivariate time-series anomaly detection.}
\label{tab:dataset-summary}
\begin{tabular}{lcccccccccc}
\toprule
\textbf{Dataset} & \textbf{Subsets} & \textbf{Features} & \multicolumn{2}{c}{\textbf{Avg Set Size}} & \multicolumn{2}{c}{\textbf{Anomalies}} & \multicolumn{3}{c}{\textbf{Segment Lengths}} & \textbf{Sampling} \\
\cmidrule(lr){4-5} \cmidrule(lr){6-7} \cmidrule(lr){8-10} \cmidrule(lr){11-11}
& & & Train & Test & Count & Ratio & Min & Med & Max & Period \\
\midrule
PSM  & 1   & 25 & 132{,}481 & 87{,}841  & 71  & 27.73\% & 1   & 5   & 8{,}861  & 1 min \\
SMD  & 28  & 38 & 25{,}300  & 25{,}301  & 327 & 4.16\%  & 2   & 11  & 3{,}161  & 1 min \\
SWaT & 1   & 51 & 496{,}800 & 449{,}919 & 34  & 12.02\% & 101 & 447 & 35{,}900 & 1 sec \\
\bottomrule
\end{tabular}
\end{table}

Table~\ref{tab:dataset-summary} summarizes the key characteristics of the benchmark datasets used in our experiments. These datasets differ substantially in terms of dimensionality, time series length, anomaly density, and segment structure.

\textbf{PSM} features relatively few but lengthy anomaly segments, resulting in a high anomaly ratio despite a limited number of anomalous events. In contrast, \textbf{SMD} comprises numerous short anomaly segments with a low overall anomaly density, making it sensitive to brief temporal disruptions. \textbf{SWaT} falls between these extremes, containing a moderate number of anomalies with segment lengths ranging from under 200 to tens of thousands of timestamps.

Together, these datasets span a spectrum of dense vs. sparse anomalies, short vs. long-term disruptions, and uni-source vs. multi-entity dynamics. This diversity allows comprehensive evaluation of model robustness under heterogeneous anomaly scenarios.

\subsection{Dataset Descriptions}
\label{appendix:datasets-descriptions}

\paragraph{PSM (Pooled Server Metrics)~\cite{PSM-2021}}  
The PSM dataset comprises server-side metrics collected from production servers at eBay. It includes 25 variables recorded over approximately 21 weeks, totaling 132{,}481 training points and 87{,}841 testing points. The test set contains both short-duration spikes and prolonged system-level anomalies, with an anomaly ratio of 27.73\%. The combination of high density and mixed-length segments makes PSM a challenging benchmark.

\paragraph{SMD (Server Machine Dataset)~\cite{OmniAnomaly-2019}}  
The SMD comprises univariate telemetry from 28 individual machines, each monitored with 38 variables. Every machine's data spans 10 days, with the first 5 days containing only normal behavior (training set) and the last 5 days including injected anomalies (testing set), resulting in an average anomaly rate of 4.16\%.

Following the original authors’ recommendation, we treat each machine as an independent subset and train a separate model per subset. This approach respects the natural isolation of system dynamics and failure modes across machines. For ablation studies and hyperparameter tuning, we exclusively utilize a few selected subsets, without overlap with those used for final evaluation.

\paragraph{SWaT (Secure Water Treatment)~\cite{SWaT-2016}}  
The SWaT dataset captures sensor and actuator data from a realistic water treatment plant testbed. It contains 51 variables over 11 days, with the first 7 days considered normal and the remaining 4 days including 41 attack scenarios. Around 12\% of the test points are labeled anomalous. SWaT’s closed-loop dynamics and complex multivariate dependencies make it especially suitable for evaluating the causal reasoning and interpretability capabilities of detection models.

\subsection{Dataset Sources}
\label{appendix:datasets-sources}

We use three widely adopted, publicly available MTSAD datasets. Each dataset is accessible for academic research, subject to the terms listed below:

\begin{itemize}
  \item \textbf{PSM (Pooled Server Metrics)}: Released by eBay and available at \url{https://github.com/netflix/psm-dataset}. Although no explicit license is provided, the dataset is publicly hosted and widely referenced in research.

  \item \textbf{SMD (Server Machine Dataset)}: First introduced in the OmniAnomaly paper~\cite{OmniAnomaly-2019}, the dataset is hosted at \url{https://github.com/NetManAIOps/OmniAnomaly} under the MIT License.

  \item \textbf{SWaT (Secure Water Treatment)}: Developed by iTrust, SUTD. The dataset requires a formal access request via \url{https://itrust.sutd.edu.sg/}. Use is permitted under a non-commercial academic research agreement.
\end{itemize}

\newpage
\section{Metrics}
\label{appendix:metrics}

\begin{table}[h]
\scriptsize
\centering
\caption{Comparison of evaluation metrics for time-series anomaly detection. 'Partial' denotes limited or conditional support depending on implementation details (e.g., decay functions, overlap thresholds).}
\label{tab:metric-comparison}
\begin{tabular}{lccccc}
\toprule
\textbf{Metric} & \textbf{Granularity} & \textbf{Threshold-Free} & \textbf{Temporal-Aware} & \textbf{Partial Credit} & \textbf{Imbalance-Robust} \\
\midrule
Standard F1    & Point-wise   & No    & No    & No     & Partial \\
Range-based F1 & Segment-wise & No    & Partial & Yes    & Partial \\
Affiliation F1 & Segment-wise & No    & Yes   & Yes    & Partial \\
AUC-ROC        & Point-wise   & Yes   & No    & No     & No      \\
AUC-PR         & Point-wise   & Yes   & No    & No     & Yes     \\
VUS-ROC        & Point+Time   & Yes   & Yes   & Yes    & Yes     \\
VUS-PR         & Point+Time   & Yes   & Yes   & Yes    & Yes     \\
\bottomrule
\end{tabular}
\end{table}

Table~\ref{tab:metric-comparison} summarizes the core properties of evaluation metrics commonly used in time-series anomaly detection. Each metric captures different facets of detection quality, including temporal alignment, threshold sensitivity, and tolerance to misclassification. Notably, no single metric is universally optimal: metrics vary in granularity (point- vs. segment-level), and in their capacity to reflect delay, calibration, or imbalance.

In our view, using multiple complementary metrics is essential for a fair and comprehensive assessment. Different datasets, anomaly types, and operational goals may favor different evaluation criteria. For instance, segment-based metrics may benefit coarse anomaly annotations, while ranking-based metrics offer insight into score separation. However, overreliance on permissive or post hoc-adjusted metrics may mask modeling deficiencies, especially in time-sensitive applications.

We assume that multivariate time-series anomaly detection (MTSAD) demands stricter evaluation. Point-wise and delay-sensitive metrics, such as Standard F1 and VUS, impose higher requirements for detection precision, temporal consistency, and calibration. While these metrics may be unforgiving, they serve as catalysts for improving the model itself, encouraging detection systems that are not merely explainable or reactive, but fundamentally precise.

\subsection{Standard F1 Score (Point-wise)}

The standard F1 score is defined as the harmonic mean of point-level precision and recall:
\begin{equation}
\text{Precision} = \frac{TP}{TP + FP}, \quad 
\text{Recall} = \frac{TP}{TP + FN},
\end{equation}
\begin{equation}
\text{F1} = \frac{2 \cdot \text{Precision} \cdot \text{Recall}}{\text{Precision} + \text{Recall}}. 
\end{equation}

Despite its well-known limitations, F1 remains one of the most rigorous indicators of model capability in pointwise anomaly detection. Evaluating each time step independently avoids assumptions about anomaly structure and instead forces the model to separate normal and anomalous scores clearly. This property can be harsh for weak models, but acts as a fair and objective measure for well-calibrated ones.

Temporal fragmentation and misalignment are often cited as weaknesses of F1. However, we argue that a strong model should inherently produce temporally coherent anomaly scores, without relying on the metric to compensate for structural noise. In this view, fragmentation bias is not a failure of the metric, but a meaningful signal about a model's inability to maintain temporal consistency.

F1 does require thresholding, which introduces variability across datasets. However, in controlled evaluation settings, identifying the optimal threshold gives insight into a model’s best achievable distinction and reflects the strength of score contrast between normal and abnormal patterns.

Finally, F1 is partially robust to class imbalance due to its exclusion of true negatives, though its reliability degrades in extremely sparse regimes. Overall, we consider F1 a high-precision, low-tolerance metric that is particularly suitable for isolating model quality in tightly controlled, point-level evaluations.

\subsection{Range-based F1 Score~\cite{PrecRec_for_TS-2018}}

We adopt a range-aware variant of the F1 score implemented in the TSB-AD benchmark suite~\cite{Elephant_in_the_Room-2024}, which provides a finer-grained assessment of segment-level anomaly detection. Unlike standard range-based F1 that uses binary overlap rules to match predicted and ground-truth segments, this version assigns \emph{partial credit} based on three interpretable components: existence, overlap, and cardinality.

The range-based recall is defined as:
\begin{equation}
\text{Recall}_{\text{range}} = \frac{1}{|\mathcal{R}|} \sum_{i=1}^{|\mathcal{R}|} \left[ \omega(R_i, \mathcal{P}) \cdot \text{CF}(R_i, \mathcal{P}) \right] + \alpha \cdot \text{ER}(\mathcal{R}, \mathcal{P}),    
\end{equation}

and $\text{Precision}_{\text{range}}$ is computed symmetrically by swapping the roles of predictions and ground truth (with $\alpha = 0$). Here, $\mathcal{P} = \{p_1, \dots, p_m\}$ denotes the set of predicted anomaly time steps, and $\mathcal{R} = \{R_1, \dots, R_{|\mathcal{R}|}\}$ the set of ground-truth anomaly ranges.

For each ground-truth range $R_i$, a weighted overlap score $\omega(R_i, \mathcal{P})$ measures the fraction of the range covered by predictions using a bias-adjusted coverage function. A \emph{cardinality factor} $\text{CF}$ penalizes cases where multiple predicted segments correspond to a single ground-truth anomaly, while an \emph{existence reward} $\text{ER}$ grants partial credit if any prediction intersects the range. The parameter $\alpha \in [0,1]$ balances the overlap and existence terms.

The final score is computed as the harmonic mean:
\begin{equation}
  \text{F1}_{\text{range}} = \frac{2 \cdot \text{Precision}_{\text{range}} \cdot \text{Recall}_{\text{range}}}{\text{Precision}_{\text{range}} + \text{Recall}_{\text{range}}}.  
\end{equation}
While this formulation enables smoother credit assignment and interpretable decomposition of detection quality, its reliance on hyperparameters (e.g., bias function, $\alpha$) and the aggregation procedure makes it sensitive to implementation details. Therefore, standardized parameter settings are essential for fair cross-method comparisons.

\subsection{Affiliation F1 Score~\cite{Local_Evaluation_TSAD-2022}}

The Affiliation F1 score evaluates prediction quality by assigning partial credit based on temporal proximity between predictions and ground-truth anomaly segments. Using soft decay functions $f(\cdot)$, it rewards predictions that are “close enough,” even if not perfectly aligned. 

The score is computed as:
\begin{equation}
\text{Precision}_{\text{aff}} = \frac{1}{m} \sum_{j=1}^{m} f_\text{p}\left(\min_{R_i \in \mathcal{R}} d(p_j, R_i)\right), \quad
\text{Recall}_{\text{aff}} = \frac{1}{|\mathcal{R}|} \sum_{i=1}^{|\mathcal{R}|} f_\text{r}\left(\min_{p_j \in \mathcal{P}} d(R_i, p_j)\right),    
\label{eq:affiliation_precision_recall}
\end{equation}
\begin{equation}
\text{F1}_{\text{aff}} = \frac{2 \cdot \text{Precision}_{\text{aff}} \cdot \text{Recall}_{\text{aff}}}{\text{Precision}_{\text{aff}} + \text{Recall}_{\text{aff}}},   
\label{eq:affiliation_f1}
\end{equation}

where function $d(p_j, R_i)$ computes the temporal distance between prediction $p_j$ and range $R_i$.

Though often described as temporally sensitive, Affiliation F1 tends to produce inflated scores even when predictions are loosely aligned with anomaly segments. The decay function $f(\cdot)$, which grants high credit to any prediction within a short temporal window, effectively smooths over temporal errors. As a result, predicting only a small portion of a segment or merely grazing its edge may yield disproportionately high F1.

This behavior is analogous to post hoc adjustments like point- or range-based scoring. It does not penalize imprecise or fragmented detection, and may mask deficiencies in a model’s ability to produce temporally coherent scores. While this soft matching is useful when ground-truth labels are coarse or ambiguous, we argue that it should be treated as a model-independent alignment compensator, not a primary measure of detection capability.

In our view, a high-performing model should not rely on the metric to explain away timing errors. Affiliation F1 may serve as a secondary diagnostic tool to assess tolerance to misalignment, but we do not consider it sufficient for rigorous performance evaluation.

\subsection{AUC-ROC~\cite{ROC-2006} and AUC-PR~\cite{PR_ROC-2006}}

AUC-ROC and AUC-PR measure the threshold-independent ranking quality of anomaly scores. For a continuous anomaly score function, they integrate model performance across all thresholds. 

The AUC-ROC computes the area under the TPR–FPR curve:
\begin{equation}
\text{TPR} = \frac{TP}{TP + FN}, \quad \text{FPR} = \frac{FP}{FP + TN}, \quad
\end{equation}
\begin{equation}
\text{AUC-ROC} = \int_{0}^{1} \text{TPR}(t) \, d(\text{FPR}(t)),
\end{equation}

while AUC-PR evaluates the precision–recall trade-off:
\begin{equation}
\text{AUC-PR} = \int_0^1 \text{Precision}(r) \, d(\text{Recall}(r)).
\end{equation}
Although widely adopted, these metrics ignore the critical temporal structure in time-series anomaly detection (TSAD). In particular, they are insensitive to detection continuity, localization lag, and whether anomalies are detected early, partially, or at all. As emphasized by Liu et al.~\cite{Elephant_in_the_Room-2024}, a model that correctly ranks only a single point in a long anomaly segment can still achieve a high AUC, even if it fails to detect the anomaly holistically.

Additionally, AUC-ROC is known to be unreliable in highly imbalanced settings, since the false positive rate denominator includes the large pool of true negatives. This allows naive models to achieve deceptively high AUC-ROC scores by merely avoiding false positives, without demonstrating meaningful detection. AUC-PR partially addresses this by focusing only on the positive class, making it more appropriate for sparse anomaly regimes.

Nonetheless, both AUC metrics are temporally agnostic and may obscure model behavior over time. We report them for completeness, but caution against over-reliance in TSAD tasks where the timing and continuity of detection are central.

\subsection{VUS-ROC and VUS-PR~\cite{VUS-2022}} 

Volume Under the Surface (VUS) metrics evaluate the consistency and timeliness of anomaly score rankings by integrating performance over both threshold levels and temporal tolerance windows. Given a set of allowable lags $\Omega = \{\omega_1, \omega_2, \dots\}$, where each $\omega$ defines a symmetric temporal buffer, the predicted points are matched to ground-truth anomalies if they fall within $\pm \omega$ time steps. For each $\omega$, we compute a corresponding AUC score:
\begin{equation}
\text{VUS-ROC} = \frac{1}{|\Omega|} \sum_{\omega \in \Omega} \text{AUC-ROC}_{\omega}, 
\end{equation}
\begin{equation}
\text{VUS-PR} = \frac{1}{|\Omega|} \sum_{\omega \in \Omega} \text{AUC-PR}_{\omega}.
\end{equation}
VUS metrics address two limitations of classical AUC: their insensitivity to detection delay and their lack of temporal context. By explicitly modeling detection within variable tolerance windows, VUS captures how consistently a model produces timely, high-ranking anomaly scores across a range of thresholds. This makes it a strong tool for evaluating delay-aware detection performance, especially in streaming or real-time settings.

Unlike segment-level metrics such as Affiliation F1, which apply smooth decay functions to soften the penalty for temporal misalignment, VUS does not interpolate between correct and incorrect predictions. It avoids heuristic credit assignment and instead tests whether the model can repeatedly detect anomalies within strict or relaxed time boundaries, without requiring threshold or decay tuning. This makes it a better measure of intrinsic anomaly scoring quality.

We find that Affiliation F1 and VUS serve complementary purposes: Affiliation F1 is suitable for analyzing misaligned segment predictions in the presence of noisy labels or lag-prone detection. At the same time, VUS offers a more principled, high-resolution assessment of detection precision and consistency across time and score thresholds.

\newpage
\section{Baselines}
\label{appendix:baselines}
\subsection{Implementation Details}
\label{appendix:baseline implementation}

All baseline models were implemented using official or widely adopted public repositories, with minor formatting adjustments to ensure evaluation consistency. Hyperparameters were either taken from the original papers or selected via lightweight tuning. All input features were normalized using a \texttt{StandardScaler} fitted on the training set (zero mean and unit variance), and no separate validation set was used. Each model was trained and evaluated using only the provided training/test splits. To ensure robustness, we repeated all experiments with five independent random seeds and report the average performance. As described in Section~\ref{sec: implemantation details}, we apply optimal threshold selection for each method to decouple detection performance from threshold sensitivity. The threshold search procedure is described in Appendix~\ref{appendix:thresholding}, and standard deviations across seeds are reported in Appendix~\ref{appendix:statistical singnificance}.

\subsection{Codebase of Implementation}
\label{appendix:baseline codebase}
Most baseline models, excluding OmniAnomaly and SARAD, are implemented using the unified codebase of CATCH:
\begin{itemize}
  \item CATCH (ICLR2025): \href{https://github.com/decisionintelligence/CATCH}{\texttt{github.com/decisionintelligence/CATCH}}
\end{itemize}

Below, we list the official implementation links for all baseline models used in our study:

\begin{itemize}
  \item OmniAnomaly (KDD 2019): \href{https://github.com/NetManAIOps/OmniAnomaly}{\texttt{github.com/NetManAIOps/OmniAnomaly}}
  \item AnomalyTransformer (ICLR 2021): \href{https://github.com/thuml/Anomaly-Transformer}{\texttt{github.com/thuml/Anomaly-Transformer}}
  \item PatchTST (ICLR 2022): \href{https://github.com/yuqinie98/PatchTST}{\texttt{github.com/yuqinie98/PatchTST}}
  \item TimesNet (ICLR 2022): \href{https://github.com/thuml/TimesNet}{\texttt{github.com/thuml/TimesNet}}
  \item DLinear (AAAI 2023): \href{https://github.com/honeywell21/DLinear}{\texttt{github.com/honeywell21/DLinear}}
  \item NLinear (AAAI 2023): \href{https://github.com/honeywell21/DLinear}{\texttt{github.com/honeywell21/DLinear}}
  \item DCdetector (KDD 2023): \href{https://github.com/DAMO-DI-ML/KDD2023-DCdetector}{\texttt{github.com/DAMO-DI-ML/KDD2023-DCdetector}}
  \item iTransformer (ICLR 2023): \href{https://github.com/thuml/iTransformer}{\texttt{github.com/thuml/iTransformer}}
  \item ModernTCN (ICLR 2023): \href{https://github.com/luodhhh/ModernTCN}{\texttt{github.com/luodhhh/ModernTCN}}
  \item SARAD (NeurIPS 2024): \href{https://github.com/daidahao/SARAD}{\texttt{github.com/daidahao/SARAD}}
\end{itemize}

\newpage
\section{More Details of Experiment}
\label{appendix:implementation}
\subsection{Implementation}

We trained all models using the AdamW~\cite{AdamW-2017} optimizer with default hyperparameters, as it provided stable convergence across datasets. To account for dataset-specific scale differences, we set the learning rate to $5\text{e}{-5}$ for PSM, while a higher value of $5\text{e}{-4}$ was used for both SMD and SWaT. Prior to training, all input features were standardized using a \texttt{StandardScaler} fitted on the training set, ensuring zero mean and unit variance. Since our evaluation protocol does not rely on validation-based model selection, we did not use a separate validation set. Instead, each model was trained and evaluated solely on the provided training/test splits.

\subsection{Hyperparameter Sensitivity}
\label{appendix:hyperparameter}
\begin{figure}[H]
    \centering
    \includegraphics[width=0.90\linewidth]{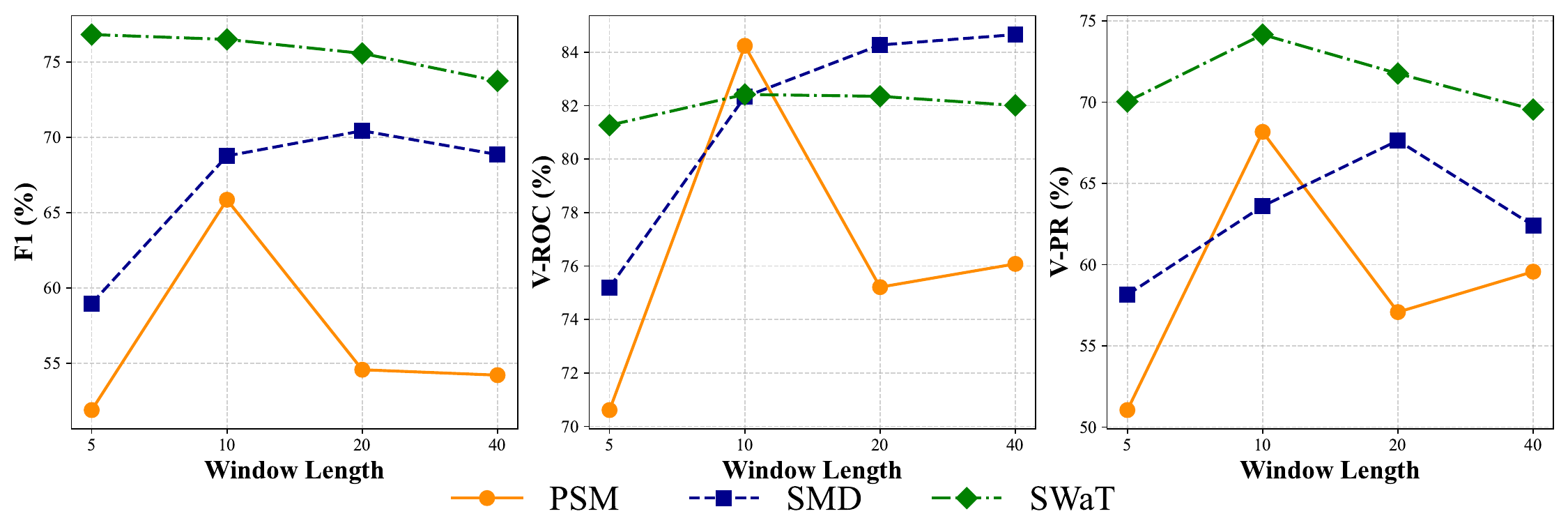}
    \caption{
Effect of input window length on OracleAD performance.  Each subplot shows a different evaluation metric. All results are averaged over five random seeds.
    }
    \label{fig:window_sensitivity}
\end{figure}

\begin{figure}[H]
    \centering
    \includegraphics[width=0.9\linewidth]{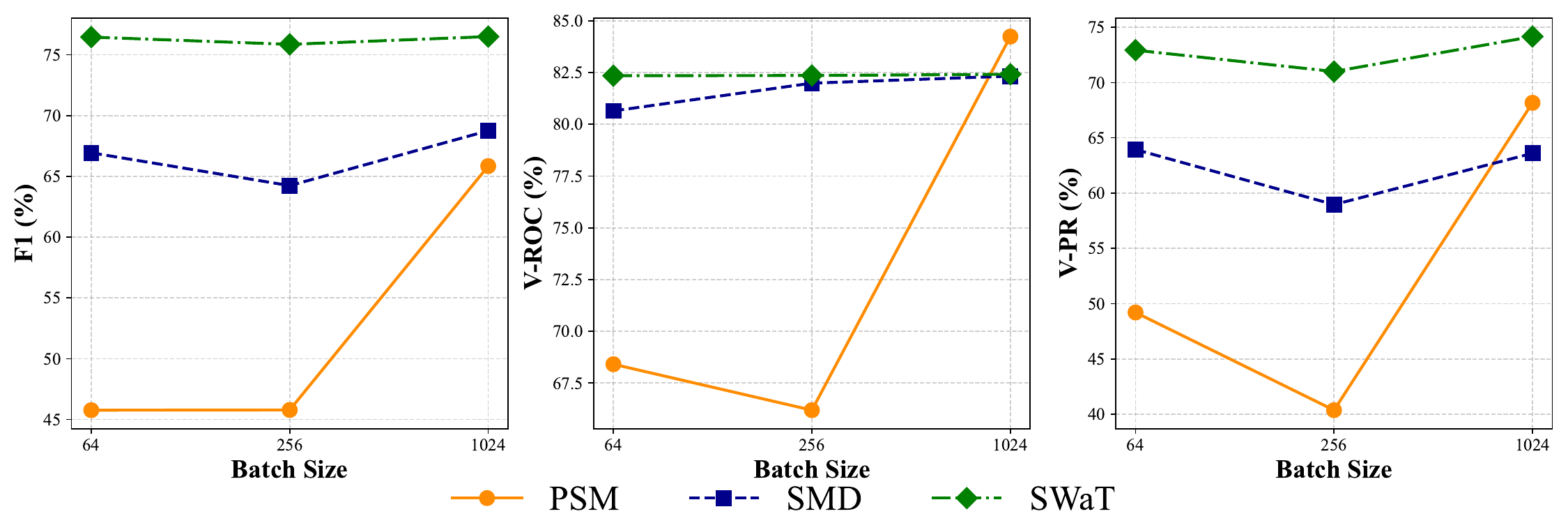}
    \caption{
Effect of training batch size on OracleAD performance. Each subplot shows a different evaluation metric. All results are averaged over five random seeds.
    }
    \label{fig:batch_sensitivity}
\end{figure}

\begin{figure}[H]
    \centering
    \includegraphics[width=0.9\linewidth]{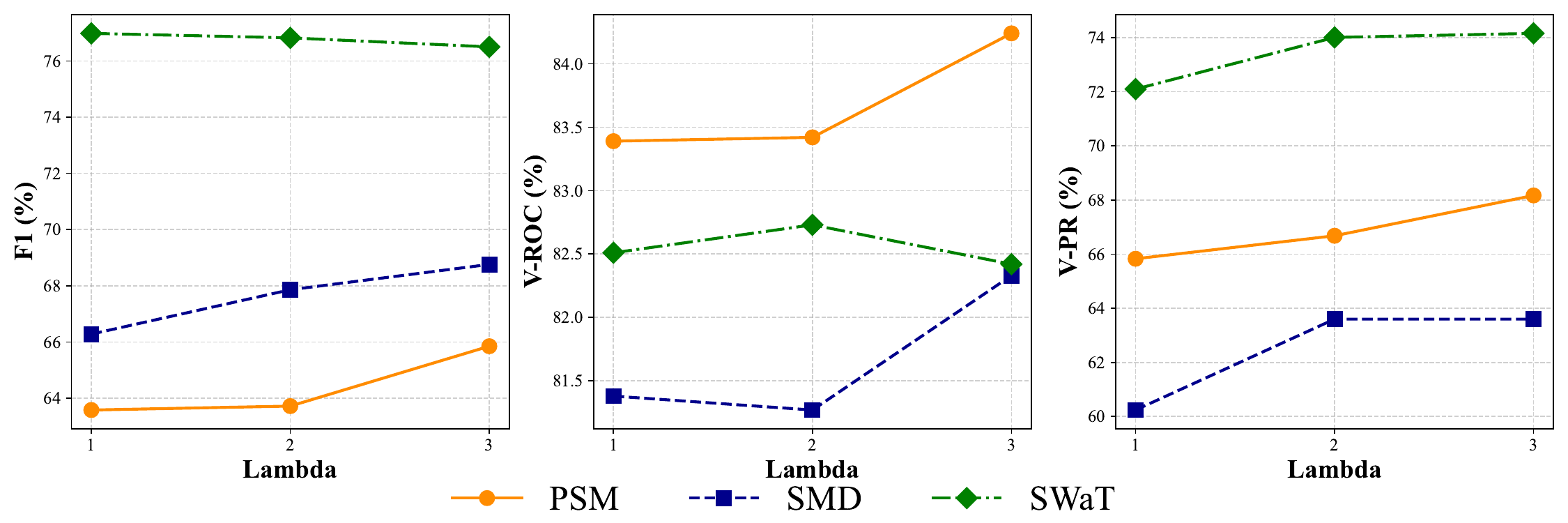}
    \caption{
Effect of the deviation regularization weight $\lambda_{\text{dev}}$ on OracleAD performance. Each subplot shows a different evaluation metric. All results are averaged over five random seeds.
    }
    \label{fig:lambda_sensitivity}
\end{figure}

\begin{figure}[H]
    \centering
    \includegraphics[width=0.9\linewidth]{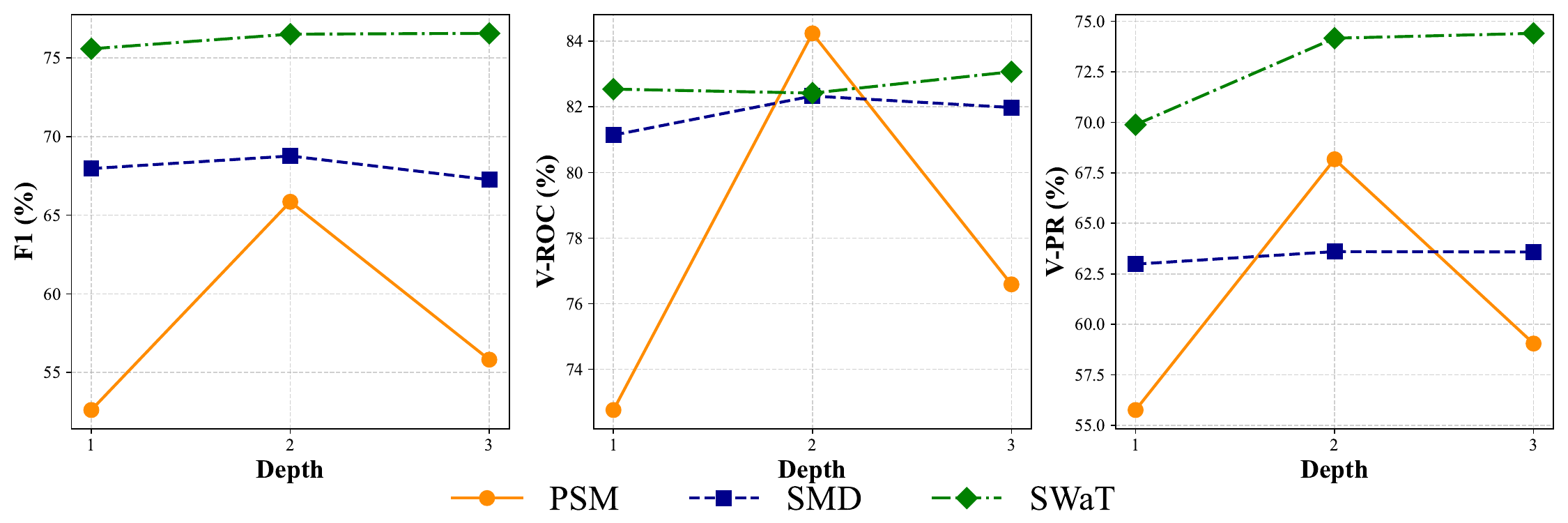}
    \caption{
Effect of  the encoder and decoder depth on OracleAD performance. Each subplot shows a different evaluation metric. All results are averaged over five random seeds.
}
    \label{fig:depth_sensitivity}
\end{figure}

We evaluated the sensitivity of OracleAD to four key hyperparameters: the window length, the batch size, the deviation weighting factor $\lambda_{\text{dev}}$, and the number of encoder and decoder layers. The analysis was performed on three benchmark datasets, and the results are summarized in Figures~\ref{fig:window_sensitivity} through \ref{fig:depth_sensitivity}, using F1, VUS-ROC, and VUS-PR as evaluation metrics.

For the SMD dataset, we followed the protocol from OmniAnomaly\footnote{\url{https://github.com/NetManAIOps/OmniAnomaly}}, treating each machine as an independent subset and selecting representative subsets for this analysis to avoid overlap with the evaluation set. The reported performance is averaged over these selected subsets.

The default configuration used across all datasets fixes the window length to 10, the batch size to 1024, the deviation weight to 3, and the number of layers to 2. While this setting does not always achieve the highest performance, it offers a stable balance between accuracy and training efficiency.

% Overall, OracleAD exhibits robust performance across a wide range of hyperparameters. Larger $\lambda_{\text{dev}}$ improves structural precision, while moderate window lengths and batch sizes yield stable results. Using more than two layers does not consistently improve performance and can reduce stability in some cases.

\paragraph{Window length.}  
As shown in Figure~\ref{fig:window_sensitivity}, OracleAD achieves its best performance with window lengths of 10 or 20 across most datasets, including PSM, SMD, and SWaT. Larger windows such as 40 occasionally improve VUS-ROC, but often lead to reduced precision in VUS-PR, likely due to the dilution of localized anomaly patterns. On the other hand, smaller windows such as 5 may fail to capture sufficient temporal context. Overall, OracleAD performs robustly within the range of 10 to 20, suggesting that this moderate window length is well suited for capturing both short-term dynamics and broader structural dependencies.

\paragraph{Batch Size.}  
Figure~\ref{fig:batch_sensitivity} shows that as batch size increases, performance improves in general. For instance, in the case of PSM, F1 score rises from 45.77 to 65.85 as batch size increases from 64 to 1024. This trend suggests that OracleAD benefits from larger batch statistics during training, potentially due to more stable gradient estimates.

\paragraph{Deviation Weight $\lambda_{\text{dev}}$.}  
As shown in Figure~\ref{fig:lambda_sensitivity}, increasing the structural regularization weight $\lambda_{\text{dev}}$ consistently improves performance across metrics. This effect is most prominent in PSM and SMD, where larger $\lambda$ values enhance VUS-ROC and VUS-PR score. The results confirm that deviation scoring based on the Stable Latent Structure plays a key role in anomaly localization and benefits from stronger guidance. In our main experiments, we set $\lambda_{\text{dev}} = 3$, as it offers the best balance between prediction and structural alignment.

\paragraph{Model Depth.}
Figure~\ref{fig:depth_sensitivity} illustrates the impact of varying the number of encoder and decoder layers. Using two layers generally leads to the best overall performance across datasets. This setting provides sufficient model capacity to capture temporal and structural dependencies without overfitting. Increasing the number of layers to three does not consistently improve performance and, in some cases such as PSM and SMD, leads to degradation in F1 score. These results suggest that additional depth may introduce unnecessary complexity or optimization instability, particularly in datasets with limited variability or high noise. Therefore, a moderate depth of two layers strikes a good balance between expressiveness and generalization.

\vspace{-1em}
\vspace{-1em}
\vspace{-1em}

\subsection{Thresholding Strategy}
\label{appendix:thresholding}

We adopt a score calibration procedure that selects the optimal threshold based on a target evaluation metric 
to convert continuous anomaly scores into binary predictions.

Given a sequence of anomaly scores $\mathbf{s} = \{s_t\}_{t=1}^T$ and corresponding binary ground-truth labels $\mathbf{y} = \{y_t\}_{t=1}^T$, the optimal threshold $\tau^*$ is obtained by maximizing an evaluation metric $\mathcal{M}$ over a discrete candidate set $\mathcal{T}$:

\begin{equation}
    \tau^* = \arg\max_{\tau \in \mathcal{T}} \ \mathcal{M}(\hat{\mathbf{y}}^\tau, \mathbf{y}),
\end{equation}

where $\hat{y}_t^\tau = \mathbb{1}[s_t \geq \tau]$ denotes the binary prediction at time $t$ based on threshold $\tau$.

The candidate set $\mathcal{T}$ is constructed by uniformly partitioning the score range into $N$ bins:

\begin{equation}
    \mathcal{T} = \left\{ \tau_i \ \middle|\ \tau_i = \min(\mathbf{s}) + \frac{i}{N - 1} \left( \max(\mathbf{s}) - \min(\mathbf{s}) \right),\ i = 0, 1, \dots, N - 1 \right\}.
\end{equation}

We set $N = 200$ in all experiments, following the protocol used by Liu et al.~\cite{Elephant_in_the_Room-2024} and adopted in the official TSB-UAD benchmark suite\footnote{\url{https://github.com/TheDatumOrg/TSB-AD}}. This value offers a good balance between threshold granularity and computational cost, and has been shown to yield stable evaluation results in large-scale studies. The threshold search is performed independently for each model and dataset to ensure fair and metric-sensitive calibration.

\subsection{System Configuration}
\label{appendix:system-configuration}

All experiments were conducted on a single workstation equipped with an NVIDIA GeForce RTX 5090 GPU (32GB) and an Intel Core Ultra 7 265K CPU (20 cores). The system had 96GB of DDR5 RAM and ran Ubuntu 24.04 LTS (64-bit). The software environment included Python 3.12, PyTorch 2.7.0, and CUDA Toolkit 12.8.

\newpage
\subsection{Computational Cost}
\label{appendix:complexity}

OracleAD employs a separate LSTM encoder–decoder for each of the $N$ variables, giving a per-window temporal cost of $O(N \cdot T)$. Multi-head self-attention requires $O(N^2)$ pairwise comparisons, leading to an overall per-window complexity of $O(N \cdot T + N^2)$.
Although this scales superlinearly with $N$, the quadratic cost is comparable to basic correlation computations and is a common requirement in models that capture explicit inter-variable structure.

We empirically profiled training and inference efficiency across real-world (PSM, SMD, SWaT) and synthetic datasets (up to $N=400$ variables). Training results (Table~\ref{tab:training-efficiency}) show that while memory usage increases with dimensionality, throughput remains practical. Inference throughput (Table~\ref{tab:inference-speed}) demonstrates that OracleAD remains feasible for real-time monitoring in systems with hundreds of variables.

\begin{table}[H]
    \caption{Training costs across datasets and batch sizes.}
    \centering
    \renewcommand{\arraystretch}{1.2}
    \setlength{\tabcolsep}{4pt}
    \begin{tabular}{c|c|c|c}
        \toprule
        Dataset (N) & Batch Size & Memory Usage & Training Speed (it/s) \\
        \midrule
        PSM (25)   & 64   & 1.5 GB  & 12.6 \\
                   & 1024 & 8.5 GB  & 9.6  \\
        \midrule
        SMD (38)   & 64   & 1.9 GB  & 8.0  \\
                   & 1024 & 13.0 GB & 6.8  \\
        \midrule
        SWaT (51)  & 64   & 2.4 GB  & 6.0  \\
                   & 1024 & 18.5 GB & 5.0  \\
        \midrule
        Syn-PSM (100) & 64  & 4.7 GB  & 3.1  \\
                     & 512 & 23.3 GB & 2.9  \\
        \midrule
        Syn-PSM (400) & 32  & 20.4 GB & 1.35 \\
        \bottomrule
    \end{tabular}
    
    \label{tab:training-efficiency}
\end{table}

\begin{table}[H]
    \caption{Inference throughput by dataset dimensionality.}
    \centering
    \renewcommand{\arraystretch}{1.2}
    \setlength{\tabcolsep}{4pt}
    \begin{tabular}{c|c}
        \toprule
        Dataset (N) & Inference Speed (time steps/s) \\
        \midrule
        PSM (25)        & 54.0 \\
        \midrule
        SMD (38)        & 37.0 \\
        \midrule
        SWaT (51)       & 28.0 \\
        \midrule
        Syn-PSM (100)   & 13.5 \\
        \midrule
        Syn-PSM (400)   & 3.15 \\
        \bottomrule
    \end{tabular}
    \label{tab:inference-speed}
\end{table}

\newpage
\section{Additional Experiments}
\label{appendix:experiments}

\subsection{Full Evaluation Results with Classical and Deep Baselines}
\label{appendix:experiments-classical}
\begin{table} [H]
\caption{Performance (\%) of anomaly detection models on the PSM, SMD, and SWaT datasets. \textbf{Boldface} indicates the best score and \underline{underline} indicates the second-best score.}
\label{tab:all_results}
\centering
\setlength{\tabcolsep}{3pt}
\renewcommand{\arraystretch}{0.95}
\begin{tiny}
\begin{tabular}{c|c|cccccccccccccccc}
\toprule
Dataset & Metric &      PCA&HBOS&IF&A.E&Omni & A.T&Patch &TsNet& DLin& NLin& DC & iTrans & Modern & SARAD & CATCH & OracleAD \\
\midrule
& F1 &      46.45&45.98&43.45&\underline{47.55}&45.90& 43.45&43.45 &43.45& 43.45& 43.45 & 43.45& 43.45& 43.45& 45.75& 44.33& \textbf{65.85}\\
 & R-F1 &      46.18&47.74&45.78&45.93&40.58& 12.17&34.75 &37.49& 34.74& 32.10& 21.35& 35.91& 35.48& 13.54& \underline{54.19}& \textbf{54.66}\\
 & Aff-F1 &      73.90&72.77&\textbf{81.98}&73.70&73.49& 69.68&69.43 &69.43& 69.43& 69.43& 69.59& 69.53& 69.43& 77.12& \underline{79.16}& 78.07\\
 PSM & A-ROC &      64.82&61.98&54.23&\underline{66.79}&63.95& 38.35&58.68 &59.09& 58.02& 58.56& 49.86& 59.22& 59.21& 62.86& 64.75& \textbf{84.78}\\
 & A-PR &      46.75&39.34&33.41&\underline{47.27}&45.00& 24.28&37.91 &38.95& 37.17& 37.64& 27.78& 38.27& 38.59& 41.67& 43.40& \textbf{68.11}\\
 & V-ROC &      66.59&63.20&56.47&\underline{68.20}&61.23& 50.96&50.62 &50.67& 50.53& 50.69& 50.11& 51.23& 50.68& 57.63& 57.00& \textbf{84.24}\\
 & V-PR &      49.13&41.18&37.78&49.66&\underline{52.49}& 49.76&50.31 &49.86& 49.93& 46.71& 27.89& 44.54& 50.55& 38.64& 45.95& \textbf{68.17}\\
\midrule
& F1 &      21.97&17.66&14.91&25.78&\underline{32.16}& 7.98&7.98 &7.98& 7.98& 7.98 & 8.00& 7.98& 7.98& 25.92& 7.98& \textbf{43.03}\\
 & R-F1 &      27.74&14.98&21.74&28.09&\underline{35.81}& 5.84&13.81 &9.43& 13.37& 13.51& 8.23& 9.02& 14.43& 10.35& 2.90& \textbf{38.88}\\
 & Aff-F1 &      72.35&69.28&81.50&\underline{83.34}&82.57& 67.43&67.43 &67.43& 67.43& 67.43 & 67.49& 67.43& 67.43& 75.87& 67.43& \textbf{84.73}\\
SMD & A-ROC &      67.88&62.57&66.44&76.80&71.25& 50.40&73.85 &59.09& 72.73& 73.83& 49.97& 74.57& 72.13& 72.84& \underline{80.96}& \textbf{83.56}\\
 & A-PR &      12.75&14.53&12.17&19.40&\underline{27.73}& 4.57&14.82 &13.66& 13.88& 14.10& 4.16& 14.57& 13.01& 25.87& 17.09& \textbf{44.83}\\
 & V-ROC &      66.50&65.24&68.58&\textbf{74.38}&\underline{73.26}& 50.34&51.66 &51.10& 51.52& 51.76& 49.92& 51.76& 51.55& 65.57& 50.95& 69.57\\
 & V-PR &      15.02&16.09&16.02&22.50&31.18& 36.86&\underline{41.60} &41.21& 41.26& 39.15& 4.27& 32.85& 41.54& 19.33& 35.25& \textbf{47.52}\\
\midrule
 & F1 &      75.36&\underline{76.04}&21.65&74.46&75.40& 21.65&21.65 &21.65& 7.98 & 12.14& 21.65& 21.65& 21.65& 57.30 & 21.65& \textbf{76.50}\\
 & R-F1 &      29.84&20.18&14.77&\underline{36.63}&\textbf{38.72}& 15.48&12.23 &20.80& 5.13 & 12.06& 15.54& 11.89& 12.88& 21.37 & 14.74& 28.15\\
 & Aff-F1 &      72.18&70.63&72.04&\textbf{75.30}&72.09& 70.06&69.51 &73.47& 67.43 & 69.68& 69.25& 69.25& 72.08& \underline{74.88} & 73.68& 71.97\\
SWaT & A-ROC &      81.79&\underline{83.52}&36.35&81.97&82.15& 42.58&24.36 &28.97& 50.48 & 23.31& 49.89& 24.57& 24.50& \textbf{85.40} & 33.31& 82.71\\
 & A-PR &      72.47&\textbf{73.99}&10.27&67.51&\underline{72.73}& 11.93&8.40 &10.71& 4.53 & 8.00& 12.14& 8.33& 9.00& 64.77 & 13.39& 72.39\\
 & V-ROC &      81.75&\underline{83.49}&41.37&81.89&81.98& 50.72&49.84 &20.80& 50.36 & 49.02& 50.00& 49.90& 50.15& \textbf{86.30} & 51.89& 82.42\\
 & V-PR &      52.31&\underline{73.90}&10.17&65.89&64.42& 17.00&11.92 &51.20& 28.47 & 10.55& 11.92& 12.05& 13.05& 62.72 & 18.70& \textbf{74.16}\\
\bottomrule
\end{tabular}
\end{tiny}
\end{table}

This section presents the complete evaluation results of OracleAD alongside both deep learning and classical anomaly detection baselines on the PSM, SMD, and SWaT datasets. While deep learning models were trained and evaluated over five random seeds to capture stochastic variation, classical baselines such as Principal Component Analysis (PCA)~\cite{PCA-2003}, Histogram-Based Outlier Score (HBOS)~\cite{HBOS-2012}, and Isolation Forest (IF)~\cite{Isolation_Forest-2008} were evaluated under a fixed random seed. This is because they are either fully deterministic or only marginally affected by random initialization, and thus produce effectively consistent outputs.

Interestingly, as shown in Table~\ref{tab:all_results}, several classical baselines outperform many deep learning models across a range of metrics. This reflects a broader phenomenon we also highlighted in Section~\ref{introduction}, complex or novel architectures do not necessarily lead to better performance. We observe that many deep models achieve their highest F1 scores when recall is maximized, meaning recall is close to one. This implies that thresholding tends to favor high precision, often at the cost of missing many true anomalies, especially those that fall within labeled anomaly segments. Consequently, these models often struggle on strict evaluation metrics such as VUS-PR and VUS-ROC, which emphasize consistent detection quality across thresholds and accurate alignment with ground-truth structure.

In contrast, classical models tend to produce relatively strong AUC-ROC and VUS-ROC scores. This indicates that while they have high overall detection rates, they often lack precision stability, which leads to weaker performance on precision-sensitive metrics like AUC-PR. The combination of high recall and low precision reveals that these models are often overly sensitive, generating many false positives.

Taken together, these findings clarify OracleAD’s distinctive advantage. It is explicitly designed to address both detection coverage and accuracy through a principled dual-scoring mechanism that combines prediction error with structural deviation. Unlike many deep models that overfit to a single scoring heuristic, OracleAD integrates temporal and inter-variable signals, resulting in stable performance across a broad range of evaluation criteria. This is particularly evident in metrics such as VUS-PR and VUS-ROC, where OracleAD consistently outperforms both classical and deep learning baselines.
\newpage

\subsection{Ablation study on the SMD dataset}
\label{appendix:experiments-smd}
\begin{table}[H]
\caption{Ablation results (\%) on selected SMD subsets. All results are averaged over five random seeds.}
\label{tab:ablation_vroc_vpr_only}
\scriptsize
\makebox[\textwidth][c]{  
\setlength{\tabcolsep}{3.7pt}
\begin{tabular}{c  |c |ccc|ccc|ccc|ccc}
  \toprule
  \multirow{2}{*}{\textbf{Component}} & \multirow{2}{*}{\textbf{Variant}} 
     &  \multicolumn{3}{c|}{machine 1-5}&  \multicolumn{3}{c|}{machine 2-3}&  \multicolumn{3}{c|}{machine 2-7}&  \multicolumn{3}{c}{machine 3-6}\\
  \cmidrule(lr){3-5} \cmidrule(lr){6-8} \cmidrule(lr){9-11} \cmidrule(lr){12-14}
  & &  F1&V-ROC & V-PR &  F1&V-ROC & V-PR &  F1&V-ROC & V-PR &  F1&V-ROC & V-PR \\
  \midrule
  \shortstack[l]{Loss Function} & w/o Recon Loss
    &  59.61&81.82 & 49.24 &  51.94&73.70 & 55.60 &  80.07&81.33 & 76.94 &  34.25&65.51& 35.37\\
  \midrule
  \multirow{2}{*}{\shortstack[l]{Anomaly Score}} 
    & w/o $\mathcal{P}^t_\text{score}$ 
    &  23.31&72.35 & 11.13 &  50.29&80.26& 51.84&  81.50&79.14 & 62.17 &  34.18&78.23& 22.95\\
    & w/o $\mathcal{D}^t_\text{score}$ 
    &  54.32&86.25& 49.60&  52.89&86.60 & 42.86 &  82.66&86.94 & 71.66 &  46.03&89.76& 50.70\\
  \midrule
  \multicolumn{2}{c|}{\textbf{OracleAD (Full Model)}} 
    &  62.03&82.44 & 52.15 &  53.99&81.71 & 54.83 &  80.35&81.75 & 76.42 &  44.39&77.32& 43.11\\
  \bottomrule
\end{tabular}
 }
\end{table}

The SMD dataset is characterized by its diversity in anomaly types, ranging from abrupt point anomalies to gradually developing structural failures. This heterogeneity results in variable performance across scoring mechanisms depending on the subset. For instance, the prediction-based score $\mathcal{P}^t_\text{score}$ often excels when anomalies manifest as sharp temporal deviations, whereas the deviation-based score $\mathcal{D}^t_\text{score}$ is more effective when anomalies involve shifts in inter-variable structure. Table~\ref{tab:ablation_vroc_vpr_only} shows that each score alone captures certain types of anomalies well, but fails to generalize across the full spectrum. In contrast, combining both scores consistently yields more stable performance across all subsets, striking a balance between temporal precision and structural sensitivity.

The role of the reconstruction loss also varies across subsets. In machine 2-7, excluding the reconstruction objective has little impact on performance, suggesting that prediction alone suffices to model normal dynamics. However, in subsets like machine 2-3 and 3-6, removing reconstruction objective significantly degrades structural metrics such as VUS-PR and causes instability in overall detection performance. This indicates that the reconstruction loss is a form of regularization that stabilizes the latent space and improves robustness, particularly when the data contains noise or complex normal variation.

These findings confirm that OracleAD’s complete model design offers the most consistent and generalizable performance across the varied anomaly profiles present in SMD.

\subsection{Statistical Significance}
\label{appendix:statistical singnificance}
\begin{table}[H]
\caption{
Standard deviation (\%) of anomaly detection performance across seven evaluation metrics on three benchmark datasets (PSM, SMD, and SWaT). All values are computed over five random seeds for each model. Lower values indicate higher stability and consistency across runs. Metrics include F1, R-F1, Aff-F1, AUC-ROC, AUC-PR, VUS-ROC, and VUS-PR.
}

\label{tab:statistical significance}
\centering
\setlength{\tabcolsep}{3pt}
\renewcommand{\arraystretch}{0.95}
\begin{scriptsize}
\begin{tabular}{c|c|cccclcccccccc}
\toprule
Dataset & Metric &  A.E&Omni & A.T&Patch &TsNet& DLin& NLin& DC & iTrans & Modern & SARAD & CATCH & OracleAD \\
\midrule
& F1 &  0.44&1.29& 0.00&0.00 &0.00& 0.00& 0.00& 0.00& 0.00& 0.00& 1.50& 1.59& 2.98\\
 & R-F1 &  0.48&2.49& 7.50&0.44 &0.61& 0.61& 0.11& 24.03& 1.67& 0.38& 7.54& 5.46& 2.93\\
 & Aff-F1 &  0.24&0.65& 0.55&0.00 &0.00& 0.00& 0.00& 0.36& 0.23& 0.00& 4.30& 4.76& 0.94\\
 PSM & A-ROC &  1.07&1.89& 2.10&0.06 &0.56& 0.07& 0.05& 0.29& 0.11& 0.14& 1.52& 0.78& 2.49\\
 & A-PR &  0.59&2.34& 0.88&0.10 &0.46& 0.05& 0.05& 0.24& 0.11& 0.18& 1.26& 0.75& 4.30\\
 & V-ROC &  0.96&0.56& 0.95&0.05 &0.03& 0.08& 0.01& 0.56& 0.24& 0.03& 4.98& 5.14& 2.47\\
 & V-PR &  0.56&0.52& 13.09&0.76 &1.40& 0.80& 0.51& 0.82& 1.02& 0.46& 1.75& 0.87& 4.31\\
\midrule
& F1 &  1.60&0.11& 0.00
&0.00 &0.00& 0.00& 0.00& 0.04& 0.00& 0.00& 0.35& 0.00& 0.92\\
 & R-F1 &  2.48&0.26& 3.69
&0.44 &0.84& 0.48& 0.15& 5.40& 1.21& 0.18& 2.28& 0.14& 0.93\\
 & Aff-F1 &  1.28&0.29& 0.00
&0.00 &0.00& 0.00& 0.00& 0.11& 0.00& 0.00& 0.35& 0.00& 0.78\\
SMD & A-ROC &  0.98&0.07& 0.27
&0.24 &0.42& 0.11& 0.04& 0.25& 0.11& 0.15& 0.32& 0.23& 0.47\\
 & A-PR &  1.78&0.11& 0.10
&0.13 &0.32& 0.06& 0.01& 0.25& 0.18& 0.06& 2.53& 0.22& 0.90\\
 & V-ROC &  2.69&0.09& 0.21
&0.08 &0.04& 0.07& 0.04& 0.22& 0.18& 0.03& 0.75& 0.08& 0.46\\
 & V-PR &  2.20&0.10& 4.35
&0.60 &0.52& 0.47& 0.26& 0.07& 0.45& 0.34& 0.37& 0.89& 1.34\\
\midrule
 & F1 &  2.34&0.04
& 0.00
&0.00 &0.00& 0.00& 0.00& 0.00& 0.00& 0.00& 3.29& 0.00& 0.31\\
 & R-F1 &  1.67&3.48
& 14.46
&0.28 &0.40& 0.12& 0.16& 0.40& 0.11& 0.60& 9.58& 0.84& 2.72\\
 & Aff-F1 &  0.77&0.05
& 0.96
&0.40 &0.64& 0.00& 0.08& 0.00& 0.00& 0.72& 1.51& 0.94& 1.80\\
SWaT & A-ROC &  0.64&0.07
& 4.39
&0.07 &0.32& 0.12& 3.54& 0.29& 0.16& 0.08& 0.91& 0.27& 1.22\\
 & A-PR &  6.48&0.02
& 0.41
&0.02 &0.09& 0.01& 1.10& 0.03& 0.01& 0.18& 12.61& 0.35& 1.99\\
 & V-ROC &  0.67&0.11
& 0.12
&0.02 &0.10& 0.06& 0.00& 0.10& 0.08& 0.08& 0.56& 0.16& 0.90\\
 & V-PR &  7.38&3.58& 0.42&0.03 &0.76& 0.04& 0.00& 0.31& 0.20& 0.19& 9.02& 0.73& 4.38\\
\bottomrule
\end{tabular}
\end{scriptsize}
\end{table}

\subsection{Distance Metrics for Dissimilarity Matrix}
\label{appendix:distance_comparison}
\begin{table}[H]
    \caption{Performance comparison of cosine similarity, L1, and L2 distances for latent vector dissimilarity on the PSM dataset. All results are averaged over five random seeds. \textbf{Boldface} indicates the best score.}
    \centering
    \scriptsize
    \renewcommand{\arraystretch}{1.2}
    \setlength{\tabcolsep}{5pt}
    \begin{tabular}{l|ccccccc}
        \toprule
        Method & F1 & R-F1 & Aff-F1 & A-ROC & A-PR & V-ROC & V-PR \\
        \midrule
        Cosine Sim. & 46.81 & 43.68 & 78.18 & 66.46 & 48.59 & 65.83 & 48.82 \\
        L1          & 57.72 & 51.54 & 75.87 & 76.33 & 54.29 & 73.33 & 53.52 \\
        L2          & \textbf{65.85} & \textbf{54.66} & \textbf{78.07} & \textbf{84.78} & \textbf{68.11} & \textbf{84.24} & \textbf{68.17} \\
        \bottomrule
    \end{tabular}
    \label{tab:distance_metrics}
\end{table}
In Eq.~\ref{eq:dissimilarity}, we define the pairwise dissimilarity matrix using the L2 distance. 
We choose L2 over other metrics because it captures both directional mismatch and magnitude differences (latent energy variation) in the embeddings. 
To validate this choice, we empirically compared L2 against cosine similarity and L1 distance across multiple evaluation metrics. 
As shown in Table~\ref{tab:distance_metrics}, L2 consistently outperforms the alternatives, yielding higher and more stable performance across five random seeds.

\subsection{Variance Dynamics of Dissimilarity Matrices}
\label{appendix:variance}
\begin{figure}[H]
    \centering
    \includegraphics[width=0.7\linewidth]{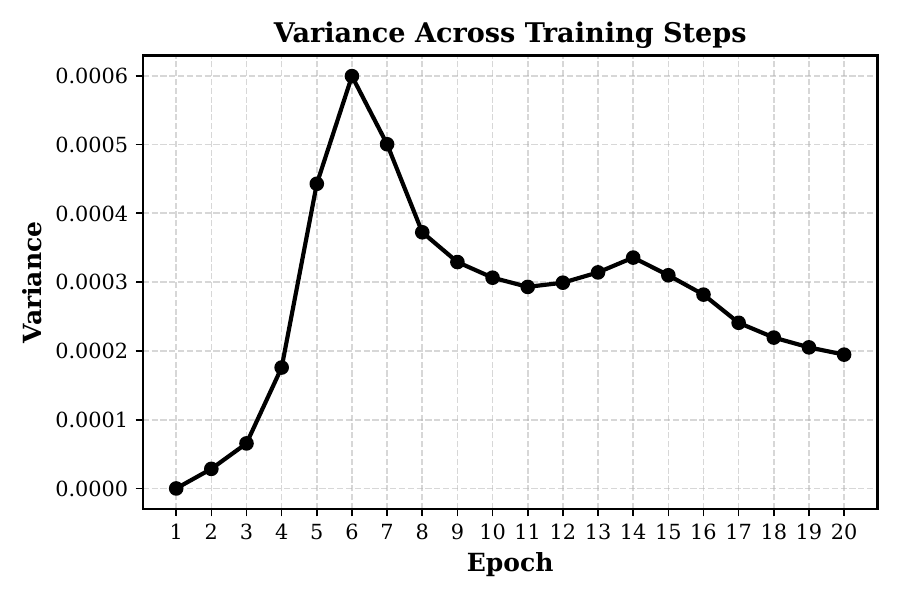}
    \caption{Variance of dissimilarity matrices across epochs on the PSM dataset. All results are averaged over five random seeds.}

    \label{fig:variance_plot}
\end{figure}
To further analyze the stability of the Stable Latent Structure (SLS), we track the variance of the dissimilarity matrices $D^{(k)}$ across training epochs. This variance reflects how much the learned inter-variable relationships fluctuate from one sliding window to another within the same epoch.

Figure~\ref{fig:variance_plot} shows the variance trend on the PSM dataset. The trajectory can be divided into two phases: in the early epochs, variance increases rapidly as the model explores latent relational geometry from sliding windows, reflecting active adaptation consistent with the reviewer’s observation that rising variance indicates meaningful structural learning; in later epochs, variance gradually decreases and converges, showing that the latent representations stabilize toward a consistent inter-variable template. This stabilization confirms that the learned SLS serves as a robust baseline for normal relational structure.

\newpage

\subsection{Comparison of Anomaly Score Fusion Strategies}
\label{appendix:score_fusion}

\begin{table}[H]
    \centering
    \scriptsize
    \renewcommand{\arraystretch}{1.15}
    \setlength{\tabcolsep}{5pt}
    \caption{Performance comparison between multiplicative and additive anomaly score fusion on PSM and SWaT datasets. Results are averaged over five random seeds.}
    \label{tab:fusion_ablation}
    \begin{tabular}{l| l|c c c c c c c}
        \toprule
        Dataset & Anomaly Score & F1 & R\text{-}F1 & Aff\text{-}F1 & A\text{-}ROC & A\text{-}PR & V\text{-}ROC & V\text{-}PR \\
        \midrule
        \multirow{2}{*}{PSM}
        & $\mathcal{P}^t_\text{score}\cdot\mathcal{D}^t_\text{score}$ & 65.85 & 54.66 & 78.07 & 84.78 & 68.11 & 84.24 & 68.17 \\
        & $\mathcal{P}^t_\text{score}+\mathcal{D}^t_\text{score}$     & 68.36 & 55.48 & 78.52 & 85.03 & 67.19 & 84.82 & 67.43 \\
        \midrule
        \multirow{2}{*}{SWaT}
        & $\mathcal{P}^t_\text{score}\cdot\mathcal{D}^t_\text{score}$ & 76.50 & 28.15 & 71.97 & 82.71 & 72.39 & 82.42 & 74.16 \\
        & $\mathcal{P}^t_\text{score}+\mathcal{D}^t_\text{score}$     & 76.91 & 37.75 & 73.43 & 81.73 & 71.73 & 81.42 & 71.07 \\
        \bottomrule
    \end{tabular}
\end{table}

To find an effective anomaly score construction, we compare two fusion strategies that combine the temporal prediction score($\mathcal{P}^t_\text{score}$) and the structural deviation score($\mathcal{D}^t_\text{score}$). Their fusion determines how OracleAD balances temporal precision and structural sensitivity.

Table~\ref{tab:fusion_ablation} presents results for multiplicative and additive fusion on PSM and SWaT. While additive fusion also achieves strong performance, it requires dataset-specific tuning to balance the relative influence of $\mathcal{P}^t_\text{score}$ and $\mathcal{D}^t_\text{score}$, as their scales and sensitivities can differ across domains. In contrast, multiplicative fusion inherently balances the two by emphasizing co-occurring signals, leading to sharper and more consistent anomaly responses. Furthermore, higher A-PR and V-PR scores indicate that the multiplicative form is more effective in reducing false positives, making it better suited for real-world applications where reliability and precision are critical.

Overall, both methods achieve competitive performance, but the multiplicative design better aligns with OracleAD’s objective, providing improved stability and interpretability in multivariate detection.

\newpage
\section{Case Studies for Interpretability}
\label{appendix:case studies}
\subsection{Complex Multivariate Anomaly}
\vspace{-0.5em}
\begin{figure}[H]
    \centering
    \includegraphics[width=0.90\linewidth]{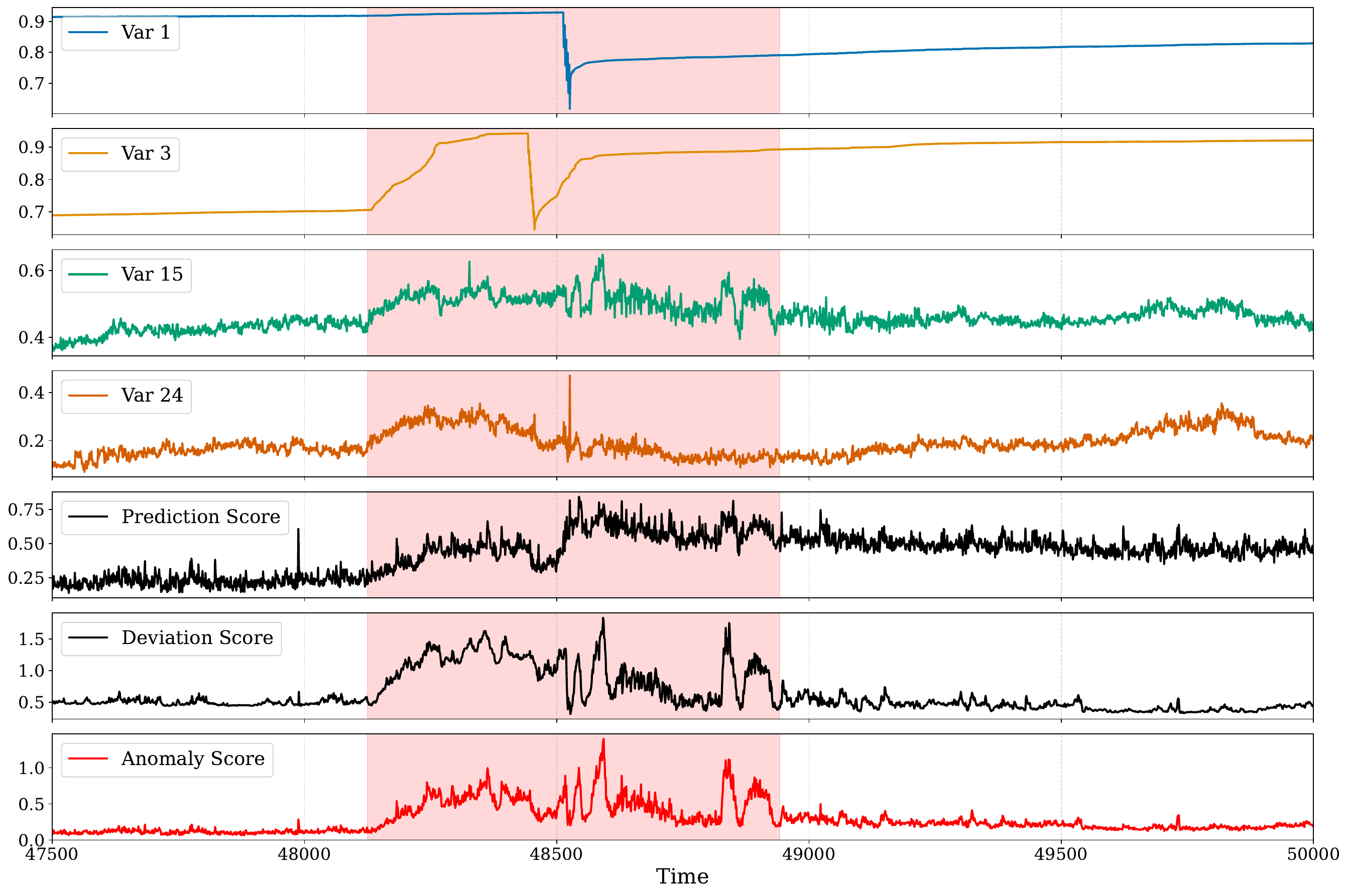}
    \caption{Raw signals of the identified root-cause variables and anomaly score visualization for selected intervals in the PSM dataset. Shaded red regions indicate ground-truth anomalies.}
    \label{fig:plot-psm}
\end{figure}
\vspace{-2em}
\begin{figure}[H]
    \centering
    \includegraphics[width=0.90\linewidth]{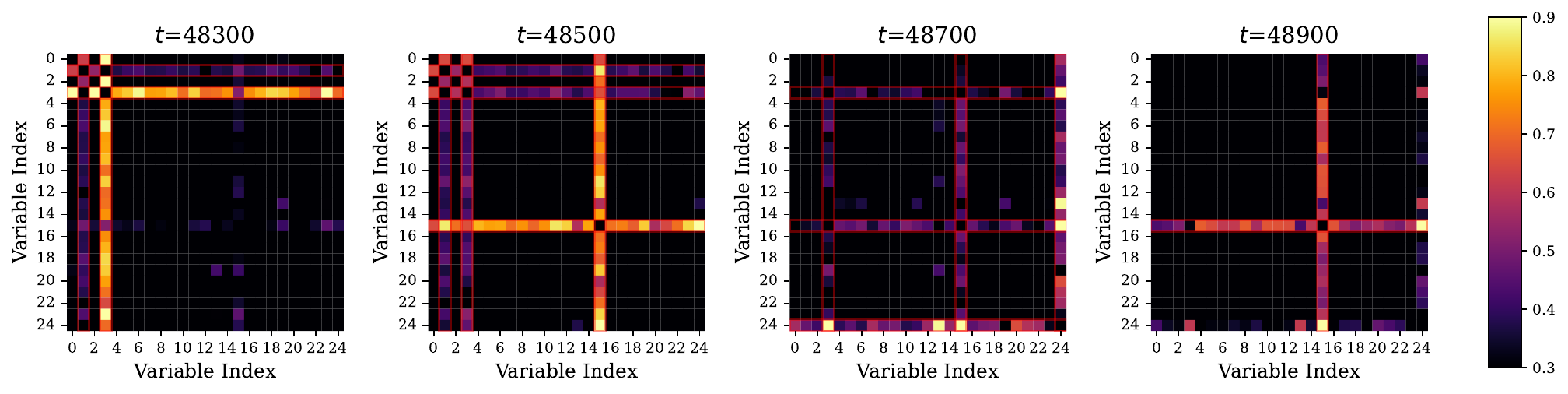}
    \caption{Visualization for Deviation Matrices $\mathcal{D}^t_\text{matrix}$ at anomalous time points in PSM dataset. Each matrix corresponds respectively to the subfigures in Figure~\ref{fig:plot-psm}. Each matrix is independently min--max normalized and values are clipped to the range [0.3, 0.9] to enhance contrast.}
    \label{fig:heatmaps-psm}
\end{figure}
Figure~\ref{fig:plot-psm} presents a case from the PSM dataset in which a single anomaly segment contains multiple overlapping anomaly signals across different variables. While the majority of variables in the PSM follow relatively stable patterns, several variables (e.g., 1, 3, 15, 24) exhibit subtle yet complex dynamics that makes it difficult to detect through prediction alone. In the highlighted anomaly segment, these variables show anomalous behavior either concurrently or sequentially, creating a challenging detection scenario.

The prediction score captures local deviations in the latter part of the segment but remains relatively low or fragmented in the earlier phase. In contrast, the deviation score maintains a more consistent profile, reflecting sustained disruptions in the latent structure. This contrast demonstrates how the deviation score is more effective in representing the internal disorganization of the system, especially when multiple variables interact in a non-trivial way.

Overall, this case illustrates that OracleAD’s dual-scoring mechanism enables anomaly detection that is both sensitive and robust across a variety of complex conditions. The prediction score is responsive to short-term deviations in individual variable behavior, while the deviation score captures structural inconsistencies across variables over time. Together, these two signals allow OracleAD to handle heterogeneous and overlapping anomaly patterns effectively.

\subsection{Long-Duration Anomaly}
\vspace{-0.5em}
\begin{figure}[H]
    \centering
    \includegraphics[width=0.90\linewidth]{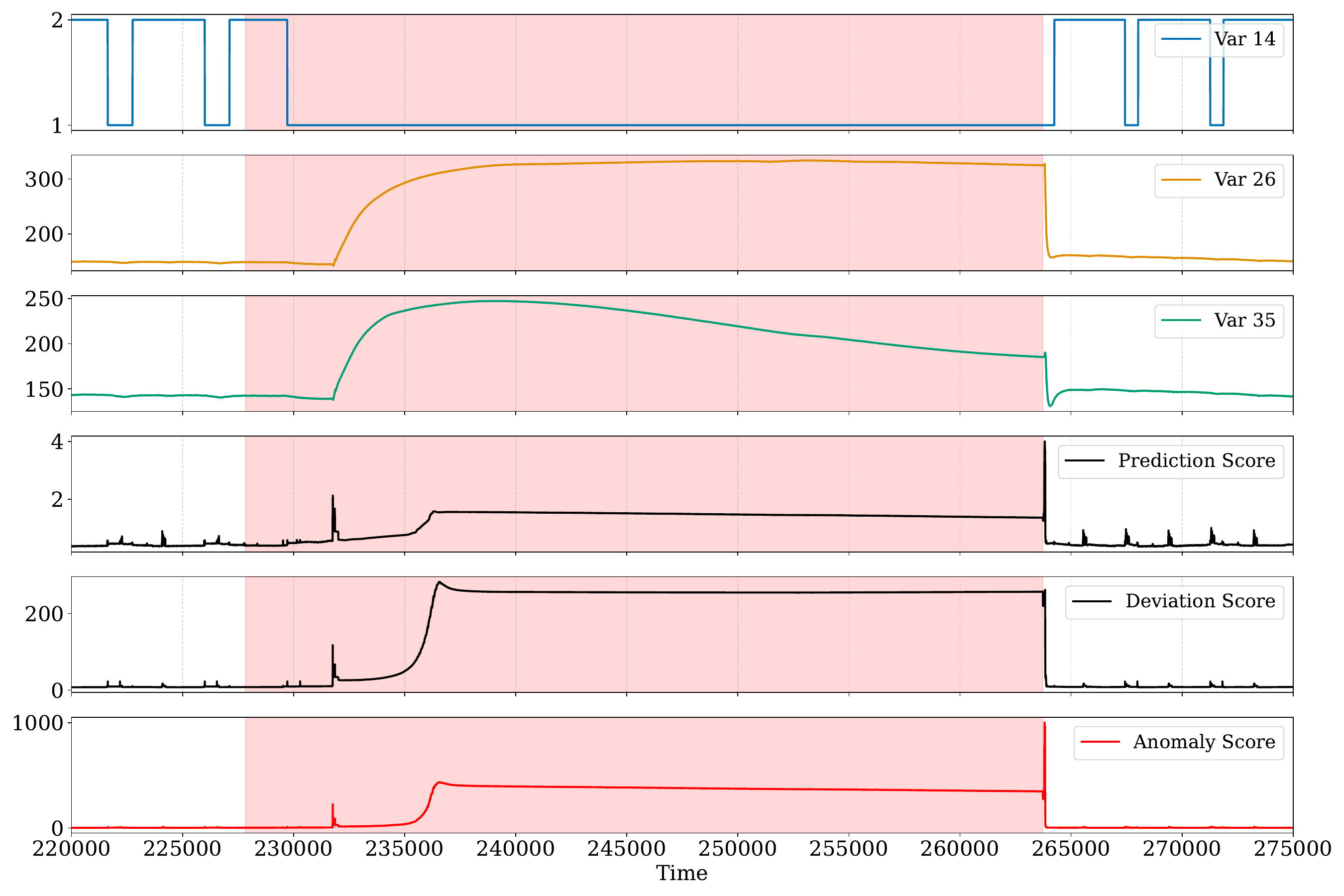}
    \caption{Raw signals of the identified root-cause variables and anomaly score visualization for selected intervals in the SWaT dataset. Shaded red regions indicate ground-truth anomalies.}
    \label{fig:plot-swat}
\end{figure}
\vspace{-2em}
\begin{figure}[H]
    \centering
    \includegraphics[width=0.90\linewidth]{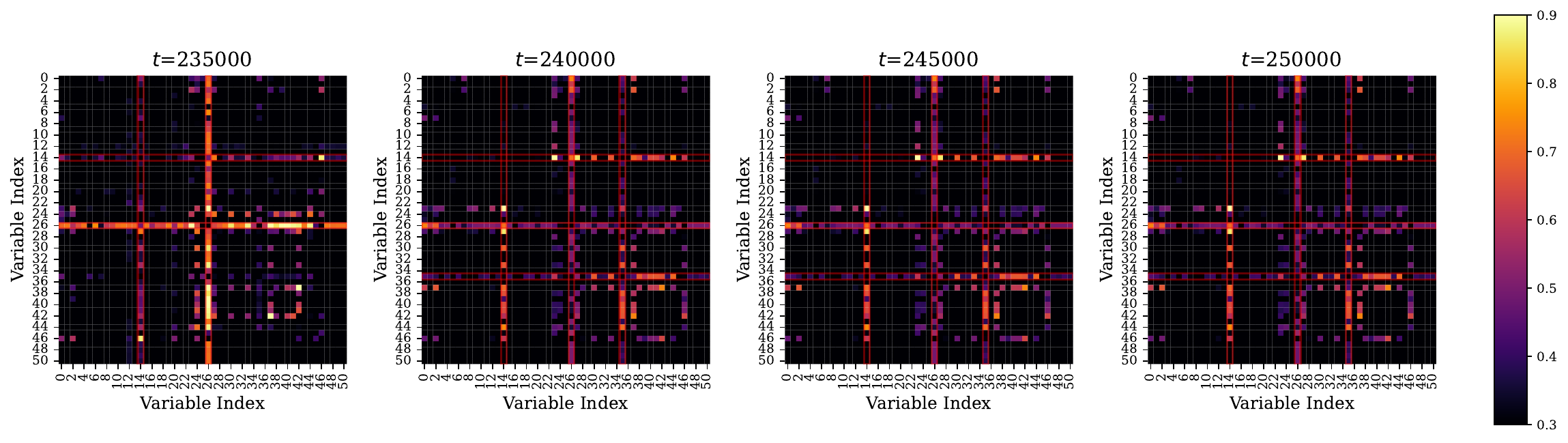}
    \caption{Visualization for Deviation Matrices $\mathcal{D}^t_\text{matrix}$ at anomalous time points in SWaT dataset. Each matrix corresponds respectively to the subfigures in Figure~\ref{fig:plot-swat}. Each matrix is independently min--max normalized and values are clipped to the range [0.3, 0.9] to enhance contrast.}
    \label{fig:heatmaps-swat}
\end{figure}

Figure~\ref{fig:plot-swat} illustrates a long anomaly segment in the SWaT dataset, notable for its persistent and structurally coherent anomaly behavior. Among all available sequences, this segment contains the longest continuous anomaly window, allowing us to evaluate OracleAD’s interpretability in a sustained abnormal context.

As shown in Figure~\ref{fig:plot-swat}, variable 26 deviates sharply and persistently from its normal operating range throughout the entire anomaly segment. This deviation is visually evident and positions variable 26 as the dominant root cause of the anomaly. Both the prediction and the deviation scores rise significantly within the anomaly interval, the deviation score plays a pivotal role in this case.

The deviation matrices, visualized in Figure~\ref{fig:heatmaps-swat}, further articulate the identification of root-cause variable. Across multiple time stamps within the anomaly segment, variable 26 consistently dominates the matrix in terms of deviation strength. This pattern indicates its sustained role as the structural root cause.

This case demonstrates how OracleAD leverages deviation scoring to detect sustained and structure-driven anomalies that may not exhibit dynamic variation within individual variables. By modeling the relational structure among variables, OracleAD maintains sensitivity even when traditional prediction-based detectors may fail due to temporal flatness.

\newpage
\section{Quantitative Analysis of Interpretability}
\begin{table}[H]
    \caption{Top-3 predicted variables ranked by aggregate deviation scores compared with ground-truth causal variables across representative anomaly timestamps in the SMD dataset. Each case is drawn from a distinct anomaly segment.}
    \centering
    \scriptsize
    \renewcommand{\arraystretch}{1.2}
    \setlength{\tabcolsep}{2pt}
    \begin{tabular}{c|cc|cc|cc|cc|cc|cc|cc|cc}
        \toprule
        & \multicolumn{2}{c|}{$t{=}2106$} & \multicolumn{2}{c|}{$t{=}4986$} &
          \multicolumn{2}{c|}{$t{=}6032$} & \multicolumn{2}{c|}{$t{=}7667$} &
          \multicolumn{2}{c|}{$t{=}14936$} & \multicolumn{2}{c|}{$t{=}15065$} &
          \multicolumn{2}{c|}{$t{=}16536$} & \multicolumn{2}{c}{$t{=}18641$} \\
        \midrule
        \multirow{4}{*}{\textbf{Top-3 Predicted}}
        & Var. & Score & Var. & Score & Var. & Score & Var. & Score
        & Var. & Score & Var. & Score & Var. & Score & Var. & Score \\
        \cmidrule(lr){2-17}
        & 13 & 9.00 & 13 & 9.49 & 32 & 15.00 & 29 & 15.99 & 33 & 12.84 & 13  & 9.39 & 13 & 13.55 & 15& 55.97\\
        & 12 & 5.46 &  8 & 7.36 & 33 & 10.16 & 33& 4.48& 32 & 6.11 &  8 & 7.50 & 12 & 6.62 & 10& 49.98\\
        & 8&  3.91& 12 & 5.79 & 31& 3.49& 34& 2.95& 35& 3.42& 12 & 5.57 &  8 & 5.56 & 18& 45.22\\
        \midrule
        \textbf{Ground Truth}
        & \multicolumn{2}{c|}{12, 13}
        & \multicolumn{2}{c|}{8, 12, 13}
        & \multicolumn{2}{c|}{32, 33}
        & \multicolumn{2}{c|}{29}
        & \multicolumn{2}{c|}{32, 33}
        & \multicolumn{2}{c|}{8, 12, 13, 14}
        & \multicolumn{2}{c|}{8, 12, 13, 14}
        & \multicolumn{2}{c}{10, 15, 18} \\
        \bottomrule
    \end{tabular}
    \label{tab:root-cause}
\end{table}
\label{appendix:quantitative analysis}
\begin{figure}[H]
    \centering
    \includegraphics[width=1.0\linewidth]{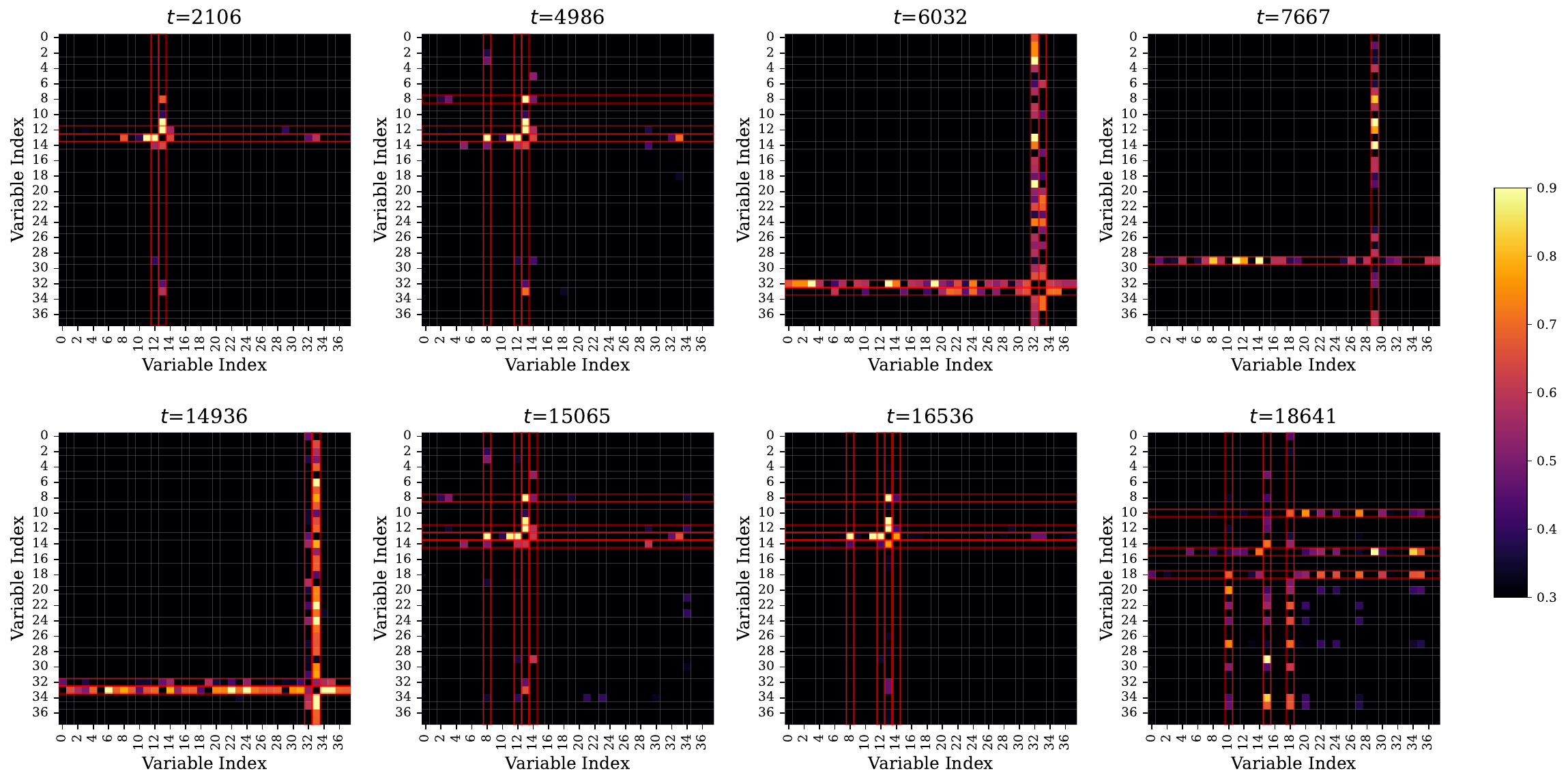}
    \caption{Deviation matrices correspond to a deviation matrix visualization in Table~\ref{tab:root-cause}. Row and column activations highlight the variables most responsible for abnormal behavior, and red boxes mark the ground-truth root-cause variables. For visualization, each matrix is independently min--max normalized and values are clipped to the range [0.3, 0.9] to enhance contrast.}
    \label{fig:heatmaps-smd-full}
\end{figure}

A distinctive strength of OracleAD lies in its ability to localize root-cause variables through the deviation matrix 
$\mathcal{D}^t_\text{matrix} = \big| D^t - \mathbf{SLS} \big|$. Each entry measures the extent to which a variable pair’s relationship deviates from the stable latent structure, and row-wise aggregation highlights which variables most strongly disrupt normal interactions.

Figure~\ref{fig:heatmaps-smd-full} shows deviation matrices at representative anomaly timestamps from the SMD dataset. Each anomaly produces a characteristic disruption pattern. As discussed in Section~\ref{experiment:diagnosis}, variables associated with multiple high-magnitude rows or columns are more likely to correspond to root causes. This enables not only root-cause analysis but also a clear visualization of how structural disruptions in key variables evolve over the course of an anomaly segment.

For precise localization, however, we rely on the aggregate deviation score obtained by summing each row (or column). This yields a variable-level anomaly score that can be ranked to produce candidate root causes. As shown in Table~\ref{tab:root-cause}, most cases quantitatively validate this approach. The top-3 ranked variables coincide with the ground-truth causal set, and even for multivariable anomalies, OracleAD reliably recovers the majority of causal variables.

\newpage
\section{Broader Impacts}

Anomaly detection in multivariate time series data has wide-reaching applications across domains such as industrial monitoring, cybersecurity, healthcare, and environmental systems. The ability to detect abnormal behavior in a timely and interpretable manner can improve operational safety, prevent financial or physical damage, and assist in early diagnostics. OracleAD aims to address key limitations of existing models by offering interpretable, structure-aware, and segment-sensitive anomaly detection.

However, as with all data-driven systems, there are potential risks. If deployed in high-stakes domains such as healthcare or autonomous control systems, undetected anomalies or false alarms may lead to significant consequences. Furthermore, the model’s performance relies heavily on the quality and representativeness of the training data. If the training data lacks diversity or contains unannotated anomalies, the model may reinforce biases or exhibit degraded performance in underrepresented conditions.

To mitigate such risks, practitioners should carefully evaluate the model under realistic operational conditions and complement automated decisions with human oversight, particularly in critical applications. We believe that OracleAD can contribute positively by promoting reliable and interpretable anomaly detection, but responsible deployment and continuous evaluation are necessary to ensure its benefits are realized without unintended harms.

\label{appendix:broader impacts}

\section{Limitations and Future Works}
\label{appendix:limiation}

While OracleAD demonstrates strong generalization and interpretability across diverse benchmarks, several limitations remain.  
First, the Stable Latent Structure (SLS) assumes globally consistent inter-variable relationships learned from continuous-valued inputs.  
This assumption may not hold in complex or multimodal systems such as SWaT, where heterogeneous subsystems or mixed variable types (e.g., continuous and discrete sensors) coexist.  
Under such conditions, the deviation matrix may lose discriminative sharpness, and the anomaly scoring can become less reliable when one component signal (prediction or deviation) weakens.

To address these challenges, future work will focus on developing (1) an anomaly detection framework that is agnostic to variable types, seamlessly handling continuous, categorical, and discrete signals; and (2) a dynamically updated SLS mechanism that adapts over time rather than remaining static during inference.  
These extensions would enhance OracleAD’s applicability to large-scale, dynamic, and heterogeneous industrial systems.

\end{document}